%% file: gedi.tex
\documentclass{article} % For LaTeX2e
\usepackage{iclr2023_conference,times}

\input{math_commands.tex}

\usepackage{amsfonts,mathtools,amssymb,amsthm}
\usepackage{wrapfig}
\usepackage[ruled]{algorithm2e}
\usepackage{algorithmic}
\usepackage{caption,subcaption}
\usepackage{booktabs,multirow}
\usepackage[colorlinks=true, hidelinks]{hyperref}
\usepackage{url}

\newcommand{\digit}[1]{\ensuremath{\vcenter{\hbox{\includegraphics[height=9pt]{img/MNIST/#1.png}}}}}

\allowdisplaybreaks

\title{Learning Symbolic Representations Through Joint GEnerative and DIscriminative Training}

\author{Emanuele Sansone \\
Department of Computer Science \\
KU Leuven\\
Leuven, 3001, Belgium \\
\texttt{emanuele.sansone@kuleuven.be} \\
\And
Robin Manhaeve \\
Department of Computer Science \\
KU Leuven\\
Leuven, 3001, Belgium \\
\texttt{robin.manhaeve@kuleuven.be} \\
}

\iclrfinalcopy % Uncomment for camera-ready version, but NOT for submission.
\begin{document}

\maketitle

\begin{abstract}
We introduce \textbf{GEDI}, a Bayesian framework that combines existing self-supervised learning objectives with likelihood-based generative models. This framework leverages the benefits of both \textbf{GE}nerative and \textbf{DI}scriminative approaches, resulting in improved symbolic representations over standalone solutions. Additionally, GEDI can be easily integrated and trained jointly with existing neuro-symbolic frameworks without the need for additional supervision or costly pre-training steps. We demonstrate through experiments on real-world data, including SVHN, CIFAR10, and CIFAR100, that GEDI outperforms existing self-supervised learning strategies in terms of clustering performance by a significant margin. % and also that it can exploit the logical contraints offered by the symbolic component to address tasks in the small data regime.
The symbolic component further allows it to leverage knowledge in the form of logical constraints to improve performance in the small data regime.
\end{abstract}

\section{Introduction}
Recently, neuro-symbolic learning has received attention as a new approach for integrating symbolic-based and sub-symbolic methods based on neural networks. This integration provides new capabilities in terms of perception and reasoning. Currently, neuro-symbolic solutions rely either on costly pre-training methods or on additional supervision at the symbolic representation level provided by the neural network, in order to effectively utilize subsequent learning feedback from the logical component~\citep{manhaeve2018deepproblog}. This traditional top-down learning paradigm is subject to the problem of \textit{representational collapse}. To gain a clearer understanding of the problem, imagine we have a tuple of three images, each of which contains a single digit (e.g., $<3, 5, 8>$). Along with this, we have information about the logical relationships between these digits (e.g., the third digit is the sum of the first two). Note that this task introduces less supervision compared to the digit addition experiment typically used in neuro-symbolic systems (i.e. the information about the sum is not provided). Current neuro-symbolic solutions can easily solve the task by mapping all input data onto the same symbol 0 and clearly solve the constrained task. 
% In this study, we present a combined approach that integrates both probabilistic logic programming and bottom-up representation learning.\rob{This claim is a bit too strong. I would rephrase it like: this}

In this study, we present a bottom-up representation learning that can naturally integrate with, and leverage the information in logical constraints.
We demonstrate that several existing self-supervised learning techniques and likelihood-based generative models can be unified within a coherent Bayesian framework called GEDI~\citep{sansone2022gedi}. The model leverages the complementary properties of discriminative approaches, which are suitable for representation learning, and of generative approaches, which capture information about the underlying density function generating the data, to learn better symbolic representations and support logical reasoning. Importantly, GEDI has two main advantages: it can be easily extended to the neuro-symbolic setting to address the collapse problem and it can also allow for learning symbolic representations in the small data regime, which is currently out of reach for existing self-supervised learning techniques.

\section{GEDI Model}
\input{src/model.tex}

%\begin{figure}[h]
%\begin{center}
%%\framebox[4.0in]{$\;$}
%\fbox{\rule[-.5cm]{0cm}{4cm} \rule[-.5cm]{4cm}{0cm}}
%\end{center}
%\caption{Sample figure caption.}
%\end{figure}

%\begin{table}[t]
%\caption{Sample table title}
%\label{sample-table}
%\begin{center}
%\begin{tabular}{ll}
%\multicolumn{1}{c}{\bf PART}  &\multicolumn{1}{c}{\bf DESCRIPTION}
%\\ \hline \\
%Dendrite         &Input terminal \\
%Axon             &Output terminal \\
%Soma             &Cell body (contains cell nucleus) \\
%\end{tabular}
%\end{center}
%\end{table}
\section{Experiments}
\input{src/experiments.tex}

%\section{Future Work}
%Our work bridges the areas of self-supervised learning and generative models, we expect to see follow-up studies proposing new implementations of the GEDI framework, such as new ways of integrating self-supervised learning and latent variable models. We have demonstrated that the GEDI solution can be easily integrated into existing statistical relational reasoning frameworks, paving the way for new neuro-symbolic integrations and enabling the handling of low data regimes, which are currently beyond the reach of existing self-supervised learning solutions.

% ---------------- After acceptance ----------------
%\subsubsection*{Author Contributions}
%\ema{TODO.}

\subsubsection*{Acknowledgments}
 This research is funded by TAILOR, a project from the EU Horizon 2020 research and innovation programme under GA No 952215. This research received also funding from the Flemish Government under the “Onderzoeksprogramma Artificiële Intelligentie (AI) Vlaanderen” programme.

\bibliography{gedi}
\bibliographystyle{iclr2023_conference}

\appendix
\input{src/appendix.tex}

\end{document}

%% file: math_commands.tex
\usepackage{amsmath,amsfonts,bm}

\def\eqref#1{equation~\ref{#1}}
\def\1{\bm{1}}

\DeclareMathAlphabet{\mathsfit}{\encodingdefault}{\sfdefault}{m}{sl}
\SetMathAlphabet{\mathsfit}{bold}{\encodingdefault}{\sfdefault}{bx}{n}

%% file: src/model.tex
\begin{wrapfigure}[21]{l}{0.4\linewidth}
     \centering
     \begin{subfigure}[b]{0.6\linewidth}
         \centering
         \includegraphics[width=0.9\textwidth]{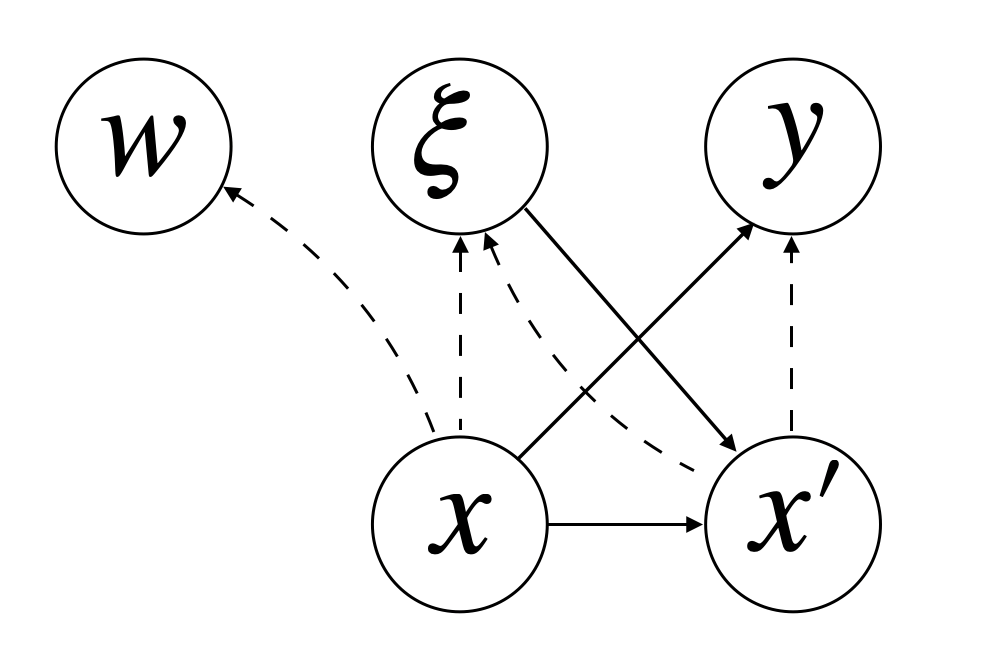}
         \caption{PGM}
     \end{subfigure}%
     \\
     \begin{subfigure}[b]{0.95\linewidth}
         \centering
         \includegraphics[width=0.9\textwidth]{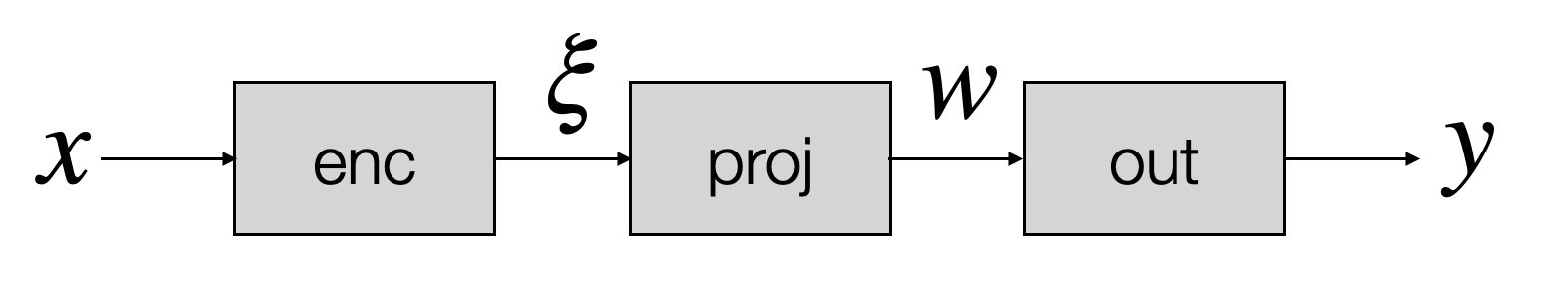}
         \caption{Block model}
     \end{subfigure}%
     \caption{GEDI model. (a) shows the corresponding probabilistic graphical model (PGM). (b) shows the different modules of GEDI, namely the encoder, the projector head and an output module computing the cosine similarity between the embedding representation and the cluster centers.}
     \label{fig:model}
\end{wrapfigure}

\textbf{Model}. Let us introduce the random quantities used in the model shown in Figure~\ref{fig:model}: (i) $x\in\Omega$, where $\Omega$ is a compact subset of $\mathbb{R}^d$, represents a data vector drawn independently from an unknown distribution $p(x)$ (for instance an image), (ii) $x'\in\Omega$ represents a transformed version of $x$ using a stochastic data augmentation strategy $\mathcal{T}(x'|x)$ (obtained by adding for instance noisy or cropping the original image) (iii) $\xi\in\mathbb{R}^h$ is the latent representation of an input data point obtained from an encoder network (the latent representation of the original image) (iv) $w\in\mathcal{S}^{h-1}$, where $\mathcal{S}^{h-1}$ is a $h-1$ dimensional unit hypersphere, is the embedding vector of an input data point (obtained from the latent representation using a network called projection head), while (v) $y\in\{1,\dots,c\}$ is the symbolic representation of an input data point defined over $c$ categories (namely the cluster label obtained by an output layer defined over the embedding representation). 
The corresponding probabilistic graphical model is given in Figure~\ref{fig:model}(a). Importantly, the generative process (solid arrows) is defined using the following conditional densities, namely: $p(w|x)=\mathcal{N}(w|0,I)$, viz. a multivariate Gaussian with zero mean and identity covariance, $p(\xi)=\mathcal{N}(\xi|0,I)$, $p(x'|x,\xi)=\mathcal{T}(x'|x)$ and $p(y|x)=\text{Softmax}(out(proj(enc(x))))$, where $enc:\Omega\rightarrow\mathbb{R}^h$ is an encoder used to compute the latent representation, $proj:\mathbb{R}^h\rightarrow\mathcal{S}^{h-1}$ is a projector head used to compute the embedding representation, and $out$ computes the cosine similarity between the embedding representation $w$ and the column vectors of a matrix of parameters $U\in\mathbb{R}^{h\times c}$ known as the cluster centers/prototypes~\citep{caron2020unsupervised}. The inference process (dashed arrows) is given by the following conditional densities: $q(w|x)=\mathcal{N}(w|0,\Sigma)$, where $\Sigma=\sum_{i=1}^n(w_i-\bar{w})(w_i-\bar{w})^T+\beta I$ is an unnormalized sample covariance matrix computed over the embedding representations of $n$ data points, $\beta$ is a positive scalar used to ensure that $\Sigma$ is positive-definite, $\bar{w}=1/n\sum_{i=1}^nw_i$ is the mean of the embedding representations; $q(\xi|enc(x)-enc(x'),I)$ assesses the level of invariance between the latent representation of the input data and its augmented version; finally $q(y|x)=\text{SK}(out(proj(enc(x'))))$ defines a distribution over cluster/prototype assignments leveraging the Sinkhorn-Knopp algorithm (SK). Please refer to the work of~\citet{caron2020unsupervised} for further details.

\textbf{Objective}. Our training objective is based on an evidence lower bound on the negative entropy, derived from the probabilistic graphical model of Figure~\ref{fig:model}(a), namely:
\begin{equation}
    E_{p(x)}\{\log p(x)\}\geq \underbrace{-CE(p,p_\Psi)}_\text{Generative term}
    + \underbrace{\mathcal{L}_{NF}(\Theta)+\mathcal{L}_{DI}(\Theta)}_\text{Self-supervised learning terms}
    \label{eq:gedi_obj}
\end{equation}
where $CE(p,p_\Psi)$ is the cross-entropy between the unknown distribution $p$ and a generative model $p_\Psi$, equivalently seen as the negative data log-likelihood of the generative model $p_\Psi$. We define $p_\Psi(x)=e^{-u^T enc(x)}/\Gamma(\Psi)$ as an energy-based model, where $\Psi$ includes both $u\in\mathbb{R}^h$ and the encoder parameters. Additionally,
\begin{align}
    \mathcal{L}_{NF}(\Theta) &= -\underbrace{\mathbb{E}_{p(x)}\{KL(q(w|x)\|p(w))\}}_\text{Decorrelation term}  -\underbrace{\mathbb{E}_{p(x)\mathcal{T}(x'|x)}\{KL(q(\xi|x,x')\|p(\xi))\}}_\text{Invariance term}
    \label{eq:nf}
\end{align}
where the first and the second addends promote decorrelated features in the embedding representation and latent representations that are invariant to data augmentations, respectively. Finally,
\begin{align}
    \mathcal{L}_{DI}(\Theta) &\geq \mathbb{E}_{p(x)\mathcal{T}(x'|x)}\{\mathbb{E}_{q(y|x')}\{\log p(y|x;\Theta)\} + H_q(y|x')\}
    \label{eq:discr}
\end{align}
where $H_q(y|x')$ is the entropy computed over $q(y|x')$ and $\Theta$ includes all parameters of the encoder, projector head and the output layer of our model. Intuitively, the first addend in Eq.~\ref{eq:discr} forces the symbolic representations of the input data and its augmented version to be similar, whereas the second addend enforces uniformity on the cluster assignments, so as to avoid that all representations collapse to a single cluster. It is important to mention that the two objectives in Eqs.~\ref{eq:nf} and~\ref{eq:discr} are general enough to cover several proposed criteria in the literature of negative-free and cluster-based self-supervised learning (cf. Appendix~\ref{sec:theory})~\citep{sansone2022gedi}. Interestingly, the objective in Eq.~\ref{eq:gedi_obj} provides a natural unification between generative and discriminative models based on self-supervised learning. Learning the GEDI model proceeds using standard gradient descent by maximizing Eq.~\ref{eq:gedi_obj} (more details about the training procedure are provided in Appendix~\ref{sec:training}).

%% file: src/experiments.tex
We perform experiments to evaluate the discriminative performance of GEDI and its competitors, namely an energy-based model JEM~\citep{grathwohl2020your}, which is trained with persistent contrastive divergence (similarly to our approach) and 2 self-supervised baselines, viz. a negative-free approach based on Barlow Twins~\citep{zbontar2021barlow} and a discriminative one based on SwAV~\citep{caron2020unsupervised}.
The whole analysis is divided into two main experimental settings, the first one based on real-world data, including SVHN, CIFAR-10 and CIFAR-100, and the second one based on a neural-symbolic learning task in the small data regime constructed from MNIST. We use existing code both as a basis to build our solution and also to run the experiments for the different baselines. In particular, we use the code from~\citet{duvenaud2021no} for training energy-based models and the repository from~\citet{costa2022solo} for all self-supervised baselines. Implementation details as well as additional experiments are reported in the Appendices.

\subsection{SVHN, CIFAR-10, CIFAR-100}\label{sec:experiments_real}
\begin{table}[t]
  \caption{Clustering performance in terms of normalized mutual information on test set (SVHN, CIFAR-10, CIFAR-100). Higher values indicate better clustering performance. We observe unstable training for SwAV on CIFAR-100. We report the best performance achieved out of 10 experiments.}
  \label{tab:nmi_real}
  \centering
\begin{tabular}{@{}lrrrrr@{}}
\toprule
\textbf{Dataset} & \textbf{JEM} & \textbf{Barlow} & \textbf{SwAV} & \textbf{GEDI} & \textbf{Gain}  \\
\midrule
SVHN & 0.04 & 0.20 & 0.24 & \textbf{0.39} & \textbf{+0.15} \\
CIFAR-10 & 0.04 & 0.22 & 0.39 & \textbf{0.41} & \textbf{+0.02}\\
CIFAR-100 & 0.05 & 0.46 & 0.69$^*$ & \textbf{0.72} & \textbf{+0.03} \\
\bottomrule
\end{tabular}
\end{table}
We consider three well-known computer vision benchmarks, namely SVHN, CIFAR-10 and CIFAR-100. We use a simple 8-layer Resnet network for the backbone encoder for both SVHN and CIFAR-10 (around 1M parameters) and increase the hidden layer size for CIFAR-100 (around 4.1M parameters) following~\citet{duvenaud2021no}. We use a MLP with a single hidden layer for $proj$ (the number of hidden neurons is twice the size of the input vector), we choose $h=256$ for CIFAR-100 and $h=128$ for all other cases. Additionally, we use data augmentation strategies commonly used in the self-supervised learning literature, including color jitter, and gray scale conversion to name a few. We train JEM, Barlow, SwAV and GEDI for $100$ epochs using Adam optimizer with learning rate $1e-4$ and batch size $64$. Further details about the hyperparameters are available in Appendix~\ref{sec:hyperparams_real}. We evaluate the clustering performance against the ground truth labels by using the Normalized Mutual Information (NMI) score.

We report all quantitative performance in Table~\ref{tab:nmi_real}. Specifically, we observe that JEM fails to solve the clustering task for all datasets. This is quite natural, as JEM is a purely generative approach, mainly designed to perform implicit density estimation. Barlow Twins achieves inferior performance to SwAV, due to the fact that is not a cluster-based self-supervised learning approach. On the contrary, we observe that GEDI is able to outperform all other competitors, thanks to the exploitation of the complementary properties of both generative and self-supervised models. Indeed, the discriminative component in GEDI leverages the information about the underlying data manifold structure learnt by the generative part, thus improving the learning of the symbolic representation. In Appendix~\ref{sec:ablation} we provide an ablation study to assess the importance of the different loss terms involved in Eq.~\ref{eq:gedi_obj}. Additionally, we conduct experiments on linear probe evaluation, generation and OOD detection tasks commonly used in the literature of self-supervised learning and energy-based models. Results are reported in Appendix~\ref{sec:additional}.

\subsection{Neural-symbolic setting}
\begin{wrapfigure}[28]{l}{0.4\linewidth}
     \centering
     \begin{subfigure}[b]{0.6\linewidth}
         \centering
         \includegraphics[width=0.9\textwidth]{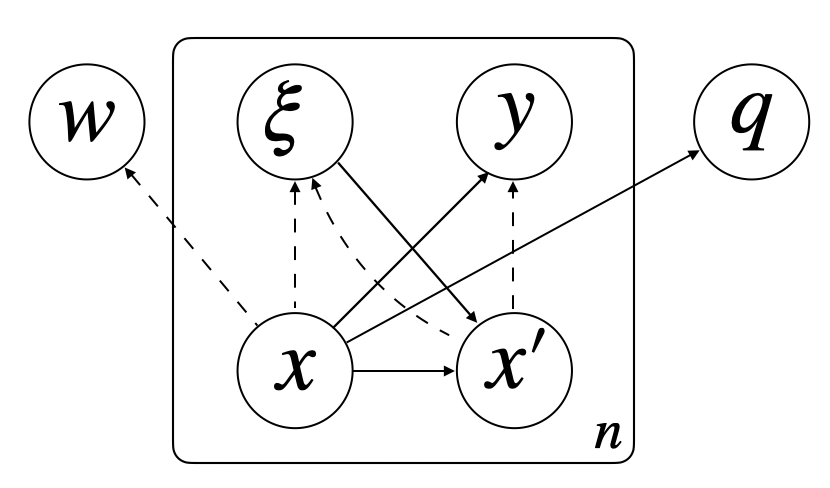}
         \caption{PGM}
     \end{subfigure}%
     \\
     \begin{subfigure}[b]{0.95\linewidth}
         \centering         \includegraphics[width=0.65\linewidth]{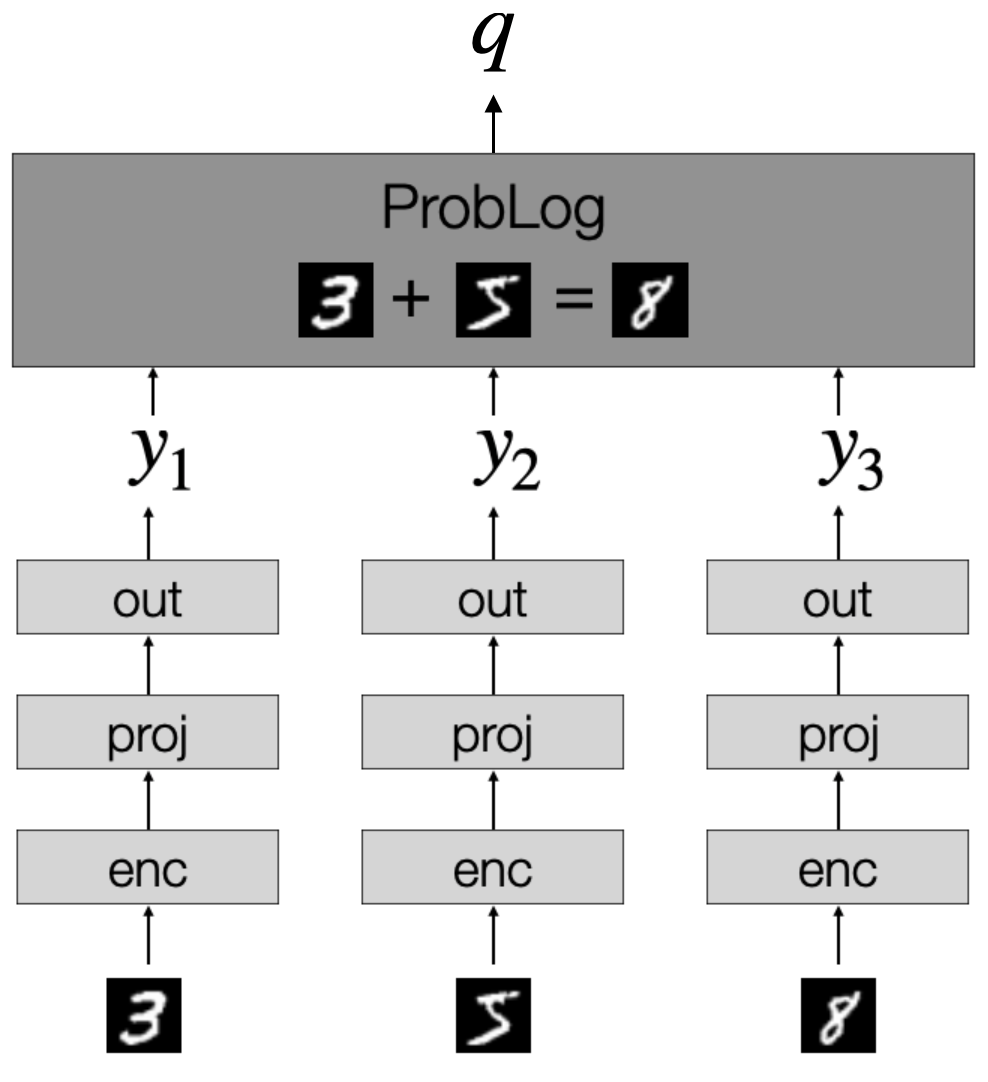}
         \caption{Block model}
     \end{subfigure}%
     \caption{Neuro-symbolic task requiring to learn the correct symbolic representation of the digits given only a tuple of images and the corresponding logical constraint. In this setting, $n=3$ data points and $q$ is a Boolean random variable used to detect if the logical constraint is satisfied.}
     \label{fig:sequence}
\end{wrapfigure}

%Task description
For the final task, we consider applying the proposed method to a neural-symbolic setting. For this, we borrow an experiment from DeepProbLog \cite{manhaeve2018deepproblog}. In this task, each example consists of a three MNIST images such that the value of the last one is the sum of the first two, e.g. $\digit{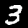} + \digit{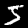} = \digit{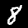}$. This can thus be considered a minimal neural-symbolic tasks, as it requires a minimal reasoning task (a single addition) on top of the image classification task. This task only contains positive examples and requires a minimal modification to the probabilistic graphical model, as shown in Figure~\ref{fig:sequence}. 
% Implementation
We use the inference mechanism from DeepProbLog to calculate the probability that this sum holds, and optimize this probability using the cross-entropy loss function, which is optimized along with the other loss functions. For this setting, this coincides with the Semantic Loss function~\citep{xu2018semantic}. To be able to calculate the probability of this addition constraint, we need the classification probabilities for each digit.

%Results
%
It is a specifically interesting use case for neural-symbolic learning, since when only the probability is optimized, the neural network tends to collapse onto the trivial solution of classifying each digit as a $0$ (i.e. $y_1=y_2=y_3=0$ in Figure~\ref{fig:sequence}). This is a logically correct but undesirable solution. Optimizing the discriminative objective should prevent this collapse. We hypothesize that a neural network can be trained to correctly classify MNIST digits by using both GEDI and the logical constraint. Since the MNIST is an easy dataset, we focus on the small data regime, and see whether the logical constraint is able to provide additional information.
%Hyperparameters
%We use ResNet~\cite{he2016resnet} for $g$ and choose $\mathcal{T}(x'|x)=N(0, \sigma^2 I)$ with $\sigma=0.1$ as our data augmentation strategy. 
The hyperparameters are identical to those used in Section~\ref{sec:experiments_real}. Further details about the hyperparameters are dependent on the data regime, and are available in Appendix~\ref{sec:hyperparams_nesy}.

We evaluate the model by measuring the accuracy and NMI of the ResNet model on the MNIST test dataset for different numbers of training examples. The results are shown in Table~\ref{tab:nesy_result}.  Here, N indicates the number of addition examples, which each have $3$ MNIST digits.
As expected, the DeepProbLog baseline from \cite{manhaeve2018deepproblog} completely fails to classify MNIST images. It has learned to map all images to the class $0$, as this results in a very low loss when considering only the logic, resultsing in an accuracy of $0.10$ and an NMI of $0.0$.
The results also show that, without the NeSy constraint, the accuracy is low for all settings. The NMI is higher, however, and increases as there is more data available. This is expected, since the model is still able to learn how to cluster from the data. However, it is  unable to correctly classify, as there is no signal in the data that is able to assign the correct label to each cluster. By including the constraint loss, the accuracy improves, as the model now has information on which cluster belongs to which class. Furthermore, it also has a positive effect on the NMI, as we have additional information on the clustering which is used by the model.
These results show us that the proposed method is beneficial to learn to correctly recognize MNIST images using only a weakly-supervised constraint, whereas other NeSy methods fail without additional regularization. Furthermore, we show that the proposed method can leverage the information offered by the constraint to further improve the NMI and classification accuracy.

\begin{table}[ht]
\caption{The accuracy and NMI of GEDI on the MNIST test set after training on the addition dataset, both with and without the NeSy constraint. Additionally, we use DeepProbLog~\citep{manhaeve2018deepproblog} as a baseline without using our GEDI model. We trained each model 5 times and report the mean and standard deviation.}
\label{tab:nesy_result}
\centering
\begin{tabular}{@{}rrrrrrr@{}}
 \toprule
 & \multicolumn{2}{c}{\textbf{Without GEDI}} & \multicolumn{2}{c}{\textbf{Without constraint}} &  \multicolumn{2}{c}{\textbf{With constraint}} \\
\textbf{N}  & 
\textbf{Acc.}               & \textbf{NMI}                &
\textbf{Acc.}               & \textbf{NMI}               &  \textbf{Acc.}             & \textbf{NMI}              \\ 
\midrule
100
& $0.10 \pm 0.00$ & $0.00 \pm 0.00$
& $0.08 \pm 0.03$ & $0.28 \pm 0.03$ & $\mathbf{0.25 \pm 0.03}$ & $\mathbf{0.41 \pm 0.03}$ \\
1000
& $0.10 \pm 0.00$ & $0.00 \pm 0.00$
& $0.09 \pm 0.02$ & $0.47 \pm 0.10$ & $\mathbf{0.52 \pm 0.26}$ & $\mathbf{0.86 \pm 0.06}$ \\
10000
& $0.10 \pm 0.00$ & $0.00 \pm 0.00$
& $0.17 \pm 0.12$ & $0.68 \pm 0.09$ & $\mathbf{0.98 \pm 0.00}$ & $\mathbf{0.97 \pm 0.01}$ \\
\bottomrule
\end{tabular}
\end{table}

%% file: src/appendix.tex
{\section{Bayesian Interpretation of Self-Supervised Learning Objectives}\label{sec:theory}}
\begin{figure*}
     \centering
     \begin{subfigure}[b]{0.24\textwidth}
         \centering \includegraphics[width=0.43\textwidth]{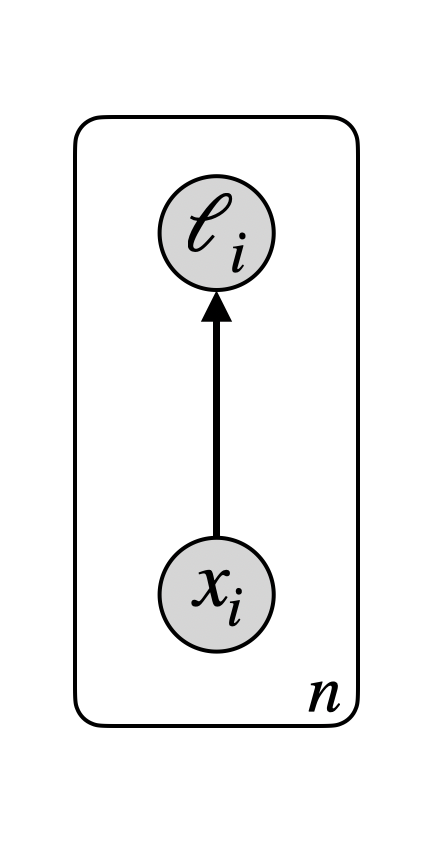}
         \caption{Contrastive (CT)}
     \end{subfigure}%   
     \begin{subfigure}[b]{0.24\textwidth}
         \centering      \includegraphics[width=0.71\textwidth]{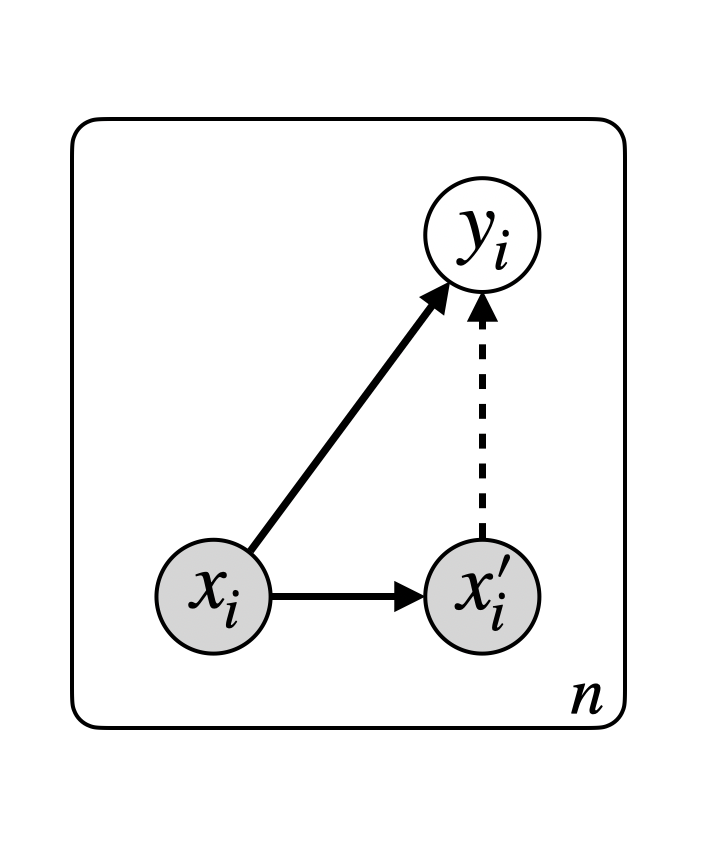}
         \caption{Discriminative (DI)}
     \end{subfigure}%
     \begin{subfigure}[b]{0.24\textwidth}
         \centering        \includegraphics[width=0.89\textwidth]{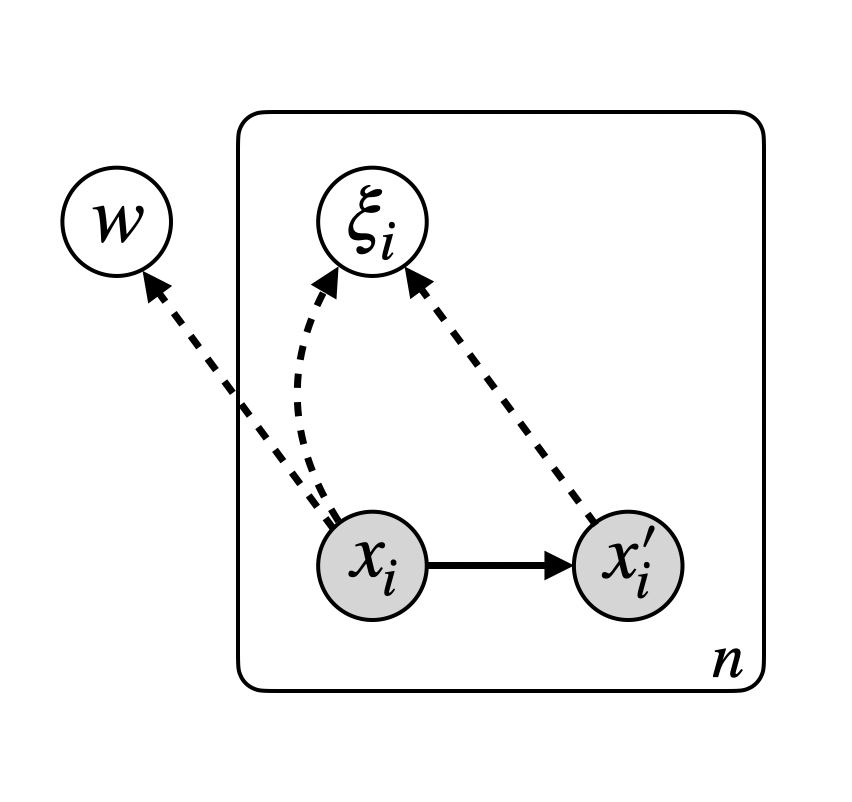}
         \caption{Negative-Free (NF)}
     \end{subfigure}%    
     %\begin{subfigure}[b]{0.24\textwidth}
        % \centering        %\includegraphics[width=0.96\textwidth]{img/PGM/Manifold.png}
         %\caption{GEDI w/o NF}
     %\end{subfigure}%     
     \begin{subfigure}[b]{0.24\textwidth}
         \centering       \includegraphics[width=1.15\textwidth]{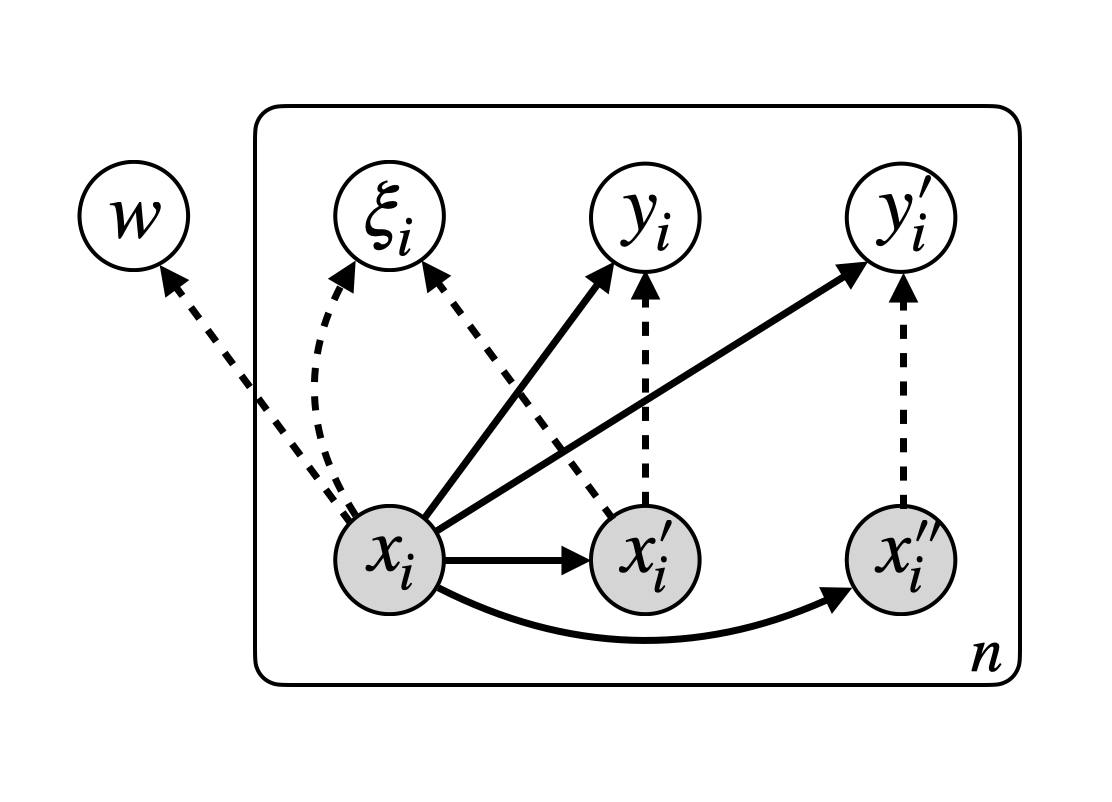}
         \caption{GEDI}
     \end{subfigure}%
     \caption{Probabilistic graphical models for the different classes of self-supervised learning approaches. White and grey nodes represent hidden 
 and observed vectors/variables, respectively. Solid arrows define the generative process, whereas dashed arrows identify auxiliary posterior densities/distributions.}
     \label{fig:graph}
\end{figure*}
We distinguish self-supervised learning approaches according to three different classes: 1) contrastive, 2) cluster-based (or discriminative) and, 3) negative-free (or non-contrastive) methods. Fig.~\ref{fig:graph} shows the corresponding probabilistic graphical models for each of the self-supervised learning classes. From Fig.~\ref{fig:graph}, we can also see how the different random quantities, we have introduced in Section 2, are probabilistically related to each other. Importantly, here we consider to have a set of input samples drawn independently from $p$ and use it $i$ to index different samples, viz. $i = \{1,\dots,n\}$

We are now ready to give an interpretation of the different SSL classes.

\subsection{Contrastive SSL}\label{sec:contrastivessl}
\input{src/contrastive}

\subsection{Discriminative/Cluster-Based SSL}
\input{src/discriminative}

\subsection{Non-contrastive/Negative-Free SSL}
\input{src/negativefree}

\section{Unifying Generative and SSL Models: A General Recipe (GEDI)}
\input{src/unified}

\begin{figure*}
    \centering    \includegraphics[width=0.7\textwidth]{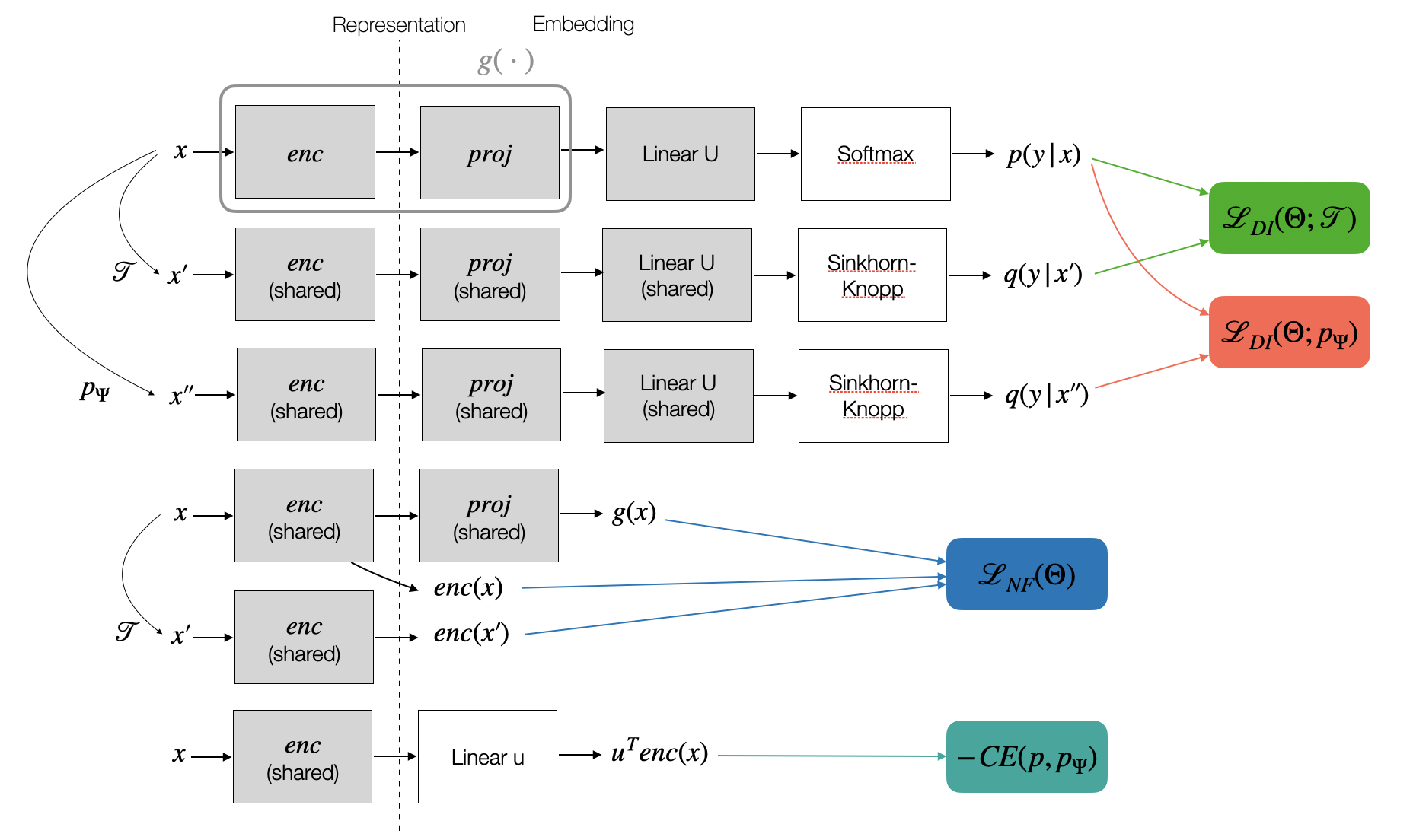}
    \caption{Diagram of the whole model. Grey boxes are shared among different rows.}
     \label{fig:wmodel}
\end{figure*}

{\section{Whole Model and Training Algorithm}\label{sec:training}}
Figure~\ref{fig:wmodel} shows the whole GEDI architecture and how to compute the different losses.
Importantly, GEDI training relies on a newly introduced data augmentation strategy, called DAM.

\begin{wrapfigure}[19]{l}{0.5\linewidth}
    \centering    \includegraphics[width=0.7\linewidth]{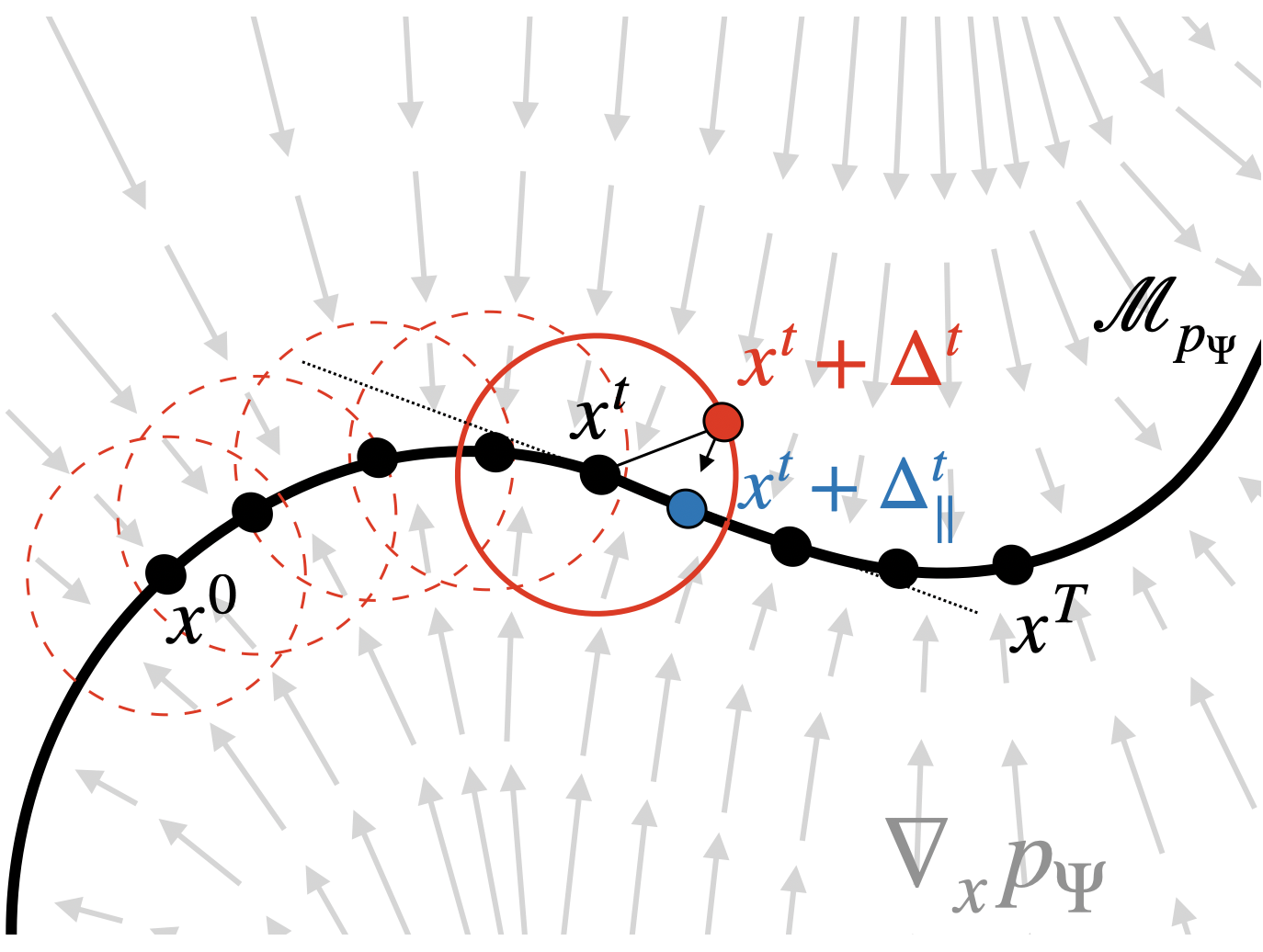}
    \caption{Data augmentation strategy exploiting the information about the manifold structure $\mathcal{M}_{p_\Psi}$ and the vector field $\nabla_xp_\Psi$ induced by the energy-based model $p_\Psi$. A local perturbation $\Delta^t$ of a point $x^t$ is projected back onto the tangent plane of the manifold by using the gradient information. The strategy is applied iteratively starting from $x\equiv x^0$ up to $x'\equiv x^T$.}
     \label{fig:data_aug}
\end{wrapfigure}

\textbf{Data Augmentation based on Manifold structure (DAM).} The routine uses generated samples obtained by "walking" on the approximated data manifold induced by the energy-based model and it assigns the same label of the original points to these samples through the discriminative loss $\mathcal{L}_{DI}(\Theta)$, thus enforcing the manifold assumption, commonly used in semi-supervised learning~\citep{chapelle2010semi}. DAM consists of an iterative procedure starting from an original training data point $x$, drawn from $p$, and generating new samples $x'$ along the approximated data manifold induced by $p_\Psi$. At each iteration $t$, the algorithm performs two main operations: Firstly, it locally perturbs a sample $x^t$ by randomly choosing a vector $\Delta^t$ on a ball of arbitrarily small radius $\epsilon>0$, viz. $B_\epsilon$, and secondly, it projects the perturbed sample $x^t+\Delta^t$ back onto the tangent plane of the approximated data manifold $\mathcal{M}_{p_\Psi}$ using the following update rule:
\begin{align}
    x^t\leftarrow x^t + \underbrace{\Delta^t -\left(\frac{\nabla_xp_\Psi(x^t+\Delta^t)^T\Delta^t}{\|\nabla_xp_\Psi(x^t+\Delta^t)\|^2}\right)\nabla_xp_\Psi(x^t+\Delta^t)}_\text{$\Delta_\parallel^t$}
    \label{eq:update_rule}
\end{align}
A visual interpretation as well as a complete description of the strategy are provided in Figure~\ref{fig:data_aug} and Algorithm~\ref{alg:dam}, respectively.
\begin{algorithm}[t]
 \caption{Data Augmentation based on Manifold structure (DAM). Formulation based on a batch of samples of size $n$.}
 \label{alg:dam}
    \textbf{Input:} $x_{1:n}, p_\Psi, \epsilon, T$\;
    \textbf{Output:} $x_{1:n}'$\;
    Sample $\Delta^0$ uniformly at random from $B_\epsilon$\;
    \textbf{For} $t=0,\dots,T$\;
    \qquad Evaluate $\nabla_xp_\Psi(x_i^t+\Delta^t)$ for all $i=1,\dots,n$\;
    \qquad Update $x_i^t$ using Eq.~(\ref{eq:update_rule}) for all $i=1,\dots,n$\;
    \qquad $\Delta^t\leftarrow \Delta^0$\;
    $x_i'\leftarrow x_i^T$ for all $i=1,\dots,n$\;
    \textbf{Return} $x_{1:n}'$\;
\end{algorithm}

\textbf{Learning a GEDI model.} The data augmentation strategy proves effective when the energy-based model approximates well the unknown data density $p(x)$. 
Consequently, we opt to train our GEDI model using a two-step procedure, where we first train the energy-based model to perform implicit density estimation and subsequently train the whole model by maximizing the objective in Eq.~(\ref{eq:gedi_obj}).
Regarding the first stage, we maximize only the generative term in Eq.~(\ref{eq:gedi_obj}) whose gradient is given by the following relation:
\begin{align}
    -\nabla_\Psi CE(p,p_\Psi) &= \mathbb{E}_{p(x)}\left\{\nabla_\Psi\log e^{u^Tenc(x)}\right\} \nonumber\\
    &\quad-\nabla_\Psi\log\Gamma(\Psi) \nonumber\\
    &=\mathbb{E}_{p(x)}\{\nabla_\Psi\log e^{u^Tenc(x)}\} \nonumber\\
    &\quad \mathbb{E}_{p_\Psi(x)}\left\{\nabla_\Psi\log e^{u^Tenc(x)}\right\}
    \label{eq:energy_obj}
\end{align}
where the first and the second expectations in Eq.~(\ref{eq:energy_obj}) are estimated using the training and the generated data, respectively. To generate data from $p_\Psi$, we use a sampler based on Stochastic Gradient Langevin Dynamics (SGLD), thus following recent best practices to train energy-based models~\citep{xie2016theory,nijkamp2019learning,du2019implicit,nijkamp2020anatomy}.

\begin{algorithm}[t]
 \caption{GEDI Training. Formulation based on a batch of samples of size $n$.}
 \label{alg:gedi}
    \textbf{Input:} $x_{1:n}$, $\text{Iters}_1$, $\text{Iters}_2$, $T$, $\epsilon$, SGLD and Adam optimizer hyperparameters\;
    \textbf{Output:} Trained model $g$\;
    \# Step 1\;
    \textbf{For} $\text{iter}=1,\dots,\text{Iters}_1$\;
    \qquad Generate samples from $p_\Psi$ using SGLD\;
    \qquad $\Psi\leftarrow\text{Adam}$ maximizing $-CE(p,p_\Psi)$\;
    \# Step 2\;
    \textbf{For} $\text{iter}=1,\dots,\text{Iters}_2$\;
    \qquad $x_{1:n}'\leftarrow\text{DAM}(x_{1:n},p_\Psi,\epsilon,T)$\;
    \qquad Generate samples from $p_\Psi$ using SGLD\;
    \qquad $\Psi,\Theta\leftarrow\text{Adam}$ maximizing Eq.~(\ref{eq:gedi_obj})\;
    \textbf{Return} $g$\;
\end{algorithm}
Regarding the second stage, we maximize the whole objective in Eq.~(\ref{eq:gedi_obj}). Specifically, at each training iteration, we run the DAM routine to obtain the augmented sample $x'$. We also use the augmented samples from the stochastic data augmentation strategy $\mathcal{T}$. Both augmentations are used to compute $\mathcal{L}_{DI}(\Theta)$ through the differentiable clustering prodecure of SwAV~\citep{caron2020unsupervised}. The learning process is summarized in Algorithm~\ref{alg:gedi}.

\textbf{Computational requirements.} Compared to traditional SSL training, and more specifically to SwAV, our learning algorithm requires additional operations and therefore increased computational requirements (but constant given $T$ and $\text{Iters}_1$). Indeed, (i) we need an additional training step (Step 1) to pre-train a generative model to approximate the unknown data density and to ensure the proper working of DAM in the second step, (ii) we need to generate samples from $p_\Psi$ to continue learning the generative model in Step 2 and (ii) we also need to run $T$ additional forward and backward passes through the energy-based model to run the DAM strategy at each iteration of the GEDI training.

{\section{Motivating Toy Examples for DAM}}
\begin{table}[t]
  \caption{Clustering performance in terms of normalized mutual information (NMI) on test set (moons and circles). Higher values indicate better clustering performance. Mean and standard deviations are computed from 5 different runs. GEDI uses $T=10$ moves in DAM.}
  \label{tab:nmi_toy}
  \centering
\begin{tabular}{@{}lrrrr@{}}
\toprule
\textbf{Dataset} & \textbf{JEM} & \textbf{SwAV} & \textbf{GEDI} & \textbf{Gain}  \\
\midrule
Moons & 0.0$\pm$0.0 & 0.8$\pm$0.1 & \textbf{0.98}$\pm$\textbf{0.02} & \textbf{+0.18}\\
Circles & 0.0$\pm$0.0 & 0.0$\pm$0.0 & \textbf{1.00}$\pm$\textbf{0.01} & \textbf{+1.00} \\
\bottomrule
\end{tabular}
\end{table}
We consider two well-known synthetic datasets, namely moons and circles. We use a multi-layer perceptron (MLP) with two hidden layers (100 neurons each) for $enc$ and one with a single hidden layer (4 neurons) for $proj$, we choose $h=2$ and we choose $\mathcal{T}(x'|x)=N(0, \sigma^2 I)$ with $\sigma=0.03$ as our data augmentation strategy. We train JEM, SwAV and GEDI for $7k$ iterations using Adam optimizer with learning rate $1e-3$. Further details about the hyperparameters are available in the  Appendix~\ref{sec:hyperparams_synth}. We evaluate the clustering performance both qualitatively, by visualizing the cluster assignments using different colors, as well as quantitatively, by using the Normalized Mutual Information (NMI) score. Furthermore, we conduct an ablation study for the different components of GEDI.

\begin{figure}
     \centering
     \begin{subfigure}[b]{0.3\linewidth}
         \centering
         \includegraphics[width=0.9\textwidth]{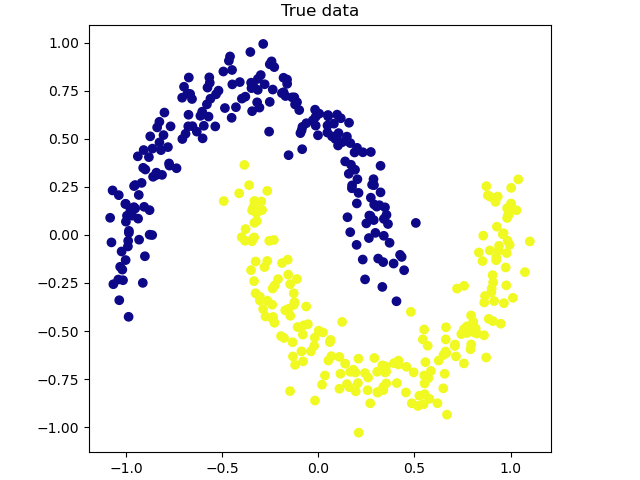}
         \caption{Ground truth}
     \end{subfigure}%
     \begin{subfigure}[b]{0.3\linewidth}
         \centering
         \includegraphics[width=0.9\textwidth]{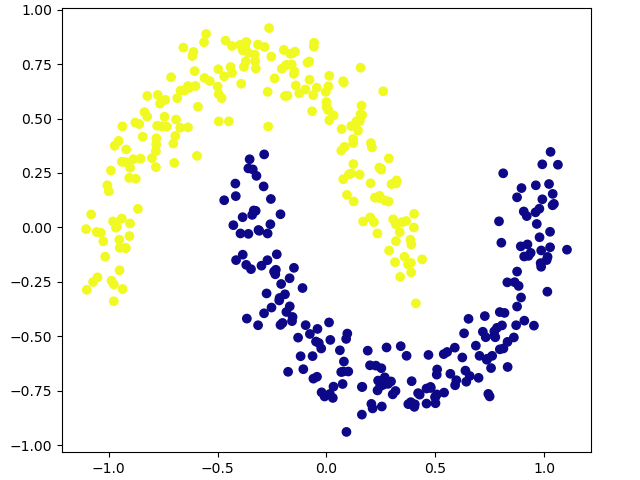}
         \caption{SwAV}
     \end{subfigure}%     
     \begin{subfigure}[b]{0.3\linewidth}
         \centering
         \includegraphics[width=0.9\textwidth]{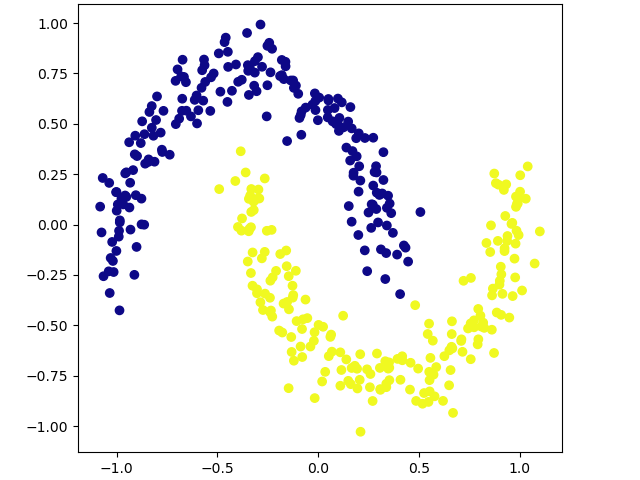}
         \caption{GEDI}
     \end{subfigure}%
     \\
    \begin{subfigure}[b]{0.3\linewidth}
         \centering
         \includegraphics[width=0.9\textwidth]{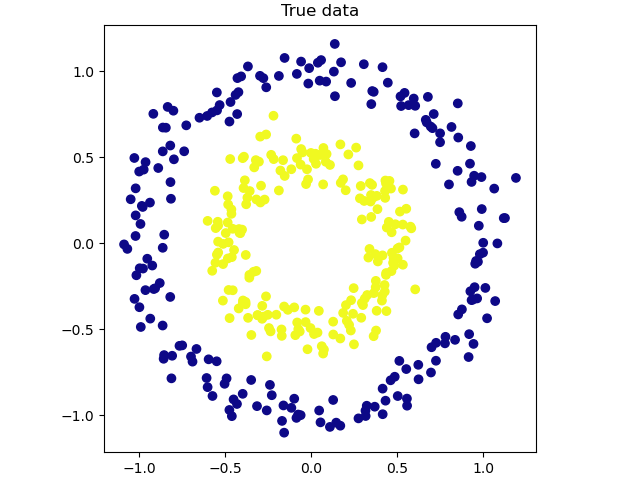}
         \caption{Ground truth}
     \end{subfigure}%
     \begin{subfigure}[b]{0.3\linewidth}
         \centering
         \includegraphics[width=0.9\textwidth]{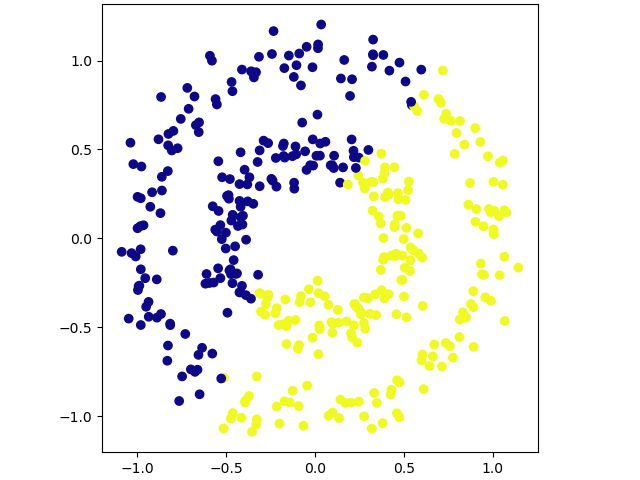}
         \caption{SwAV}
     \end{subfigure}%     
     \begin{subfigure}[b]{0.3\linewidth}
         \centering
         \includegraphics[width=0.9\textwidth]{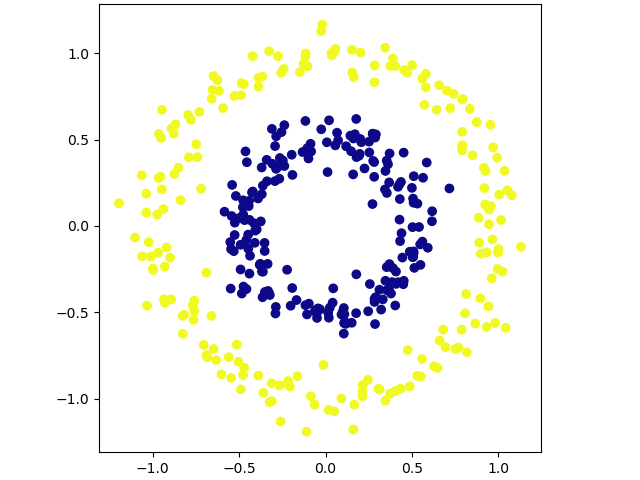}
         \caption{GEDI}
     \end{subfigure}%
     \caption{Qualitative visualization of the clustering performance for the different strategies (only SwAV and GEDI are shown) on moons (a-c) and on circles (d-f) datasets. Colors identify different cluster predictions.}
     \label{fig:labels_toy}
\end{figure}

We report all quantitative performance in Table~\ref{tab:nmi_toy}. Specifically, we observe that JEM fails to solve the clustering task for both datasets. This is quite natural, as JEM is a purely generative approach, mainly designed to perform implicit density estimation. SwAV can only solve the clustering task for the moons dataset, highlighting the fact that it is not able to exploit the information from the underlying density used to generate the data. However, GEDI can recover the true clusters in both datasets. This is due to the fact that GEDI uses the information from the generative component through DAM to inform the cluster-based one. Consequently, GEDI is able to exploit the manifold structure underlying data. Figure~\ref{fig:labels_toy} provide some examples of predictions by SwAV and GEDI on the two datasets. From the figure, we can clearly see that GEDI can recover the two data manifolds up to a permutation of the labels.

\begin{figure}
     \centering
     \begin{subfigure}[b]{0.49\linewidth}
         \centering
         \includegraphics[width=0.9\textwidth]{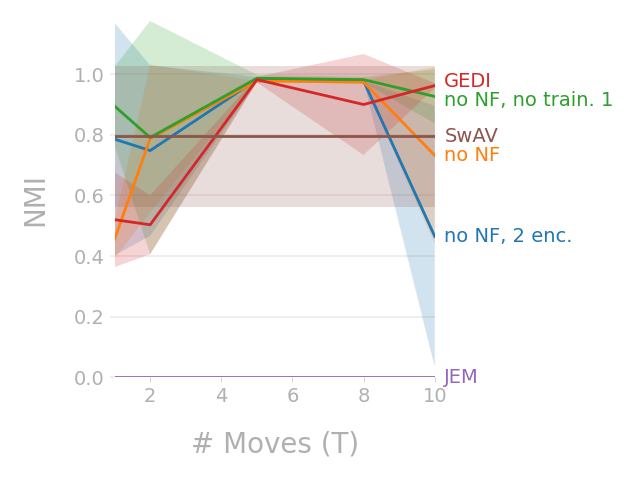}
         \caption{Moons}
     \end{subfigure}%
     \begin{subfigure}[b]{0.49\linewidth}
         \centering
         \includegraphics[width=0.9\textwidth]{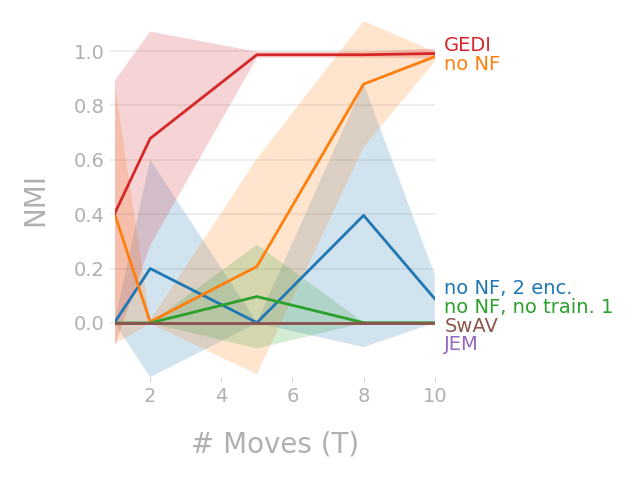}
         \caption{Circles}
     \end{subfigure}%     
     \caption{Ablation study for GEDI on moons and circles dataset. The curves are obtained by choosing different values of $T$ for DAM, namely $T\in\{1,2,5,8,10\}$.}
     \label{fig:ablation_toy}
\end{figure}

We conduct an ablation study to understand the impact of the different components of GEDI. We compare four different versions of GEDI, namely the full version (called simply \textit{GEDI}), GEDI trained without $\mathcal{L}_{NF}(\Theta)$ (called \textit{no NF}), GEDI trained without the first stage and also without $\mathcal{L}_{NF}(\Theta)$ (called \textit{no NF, no train. 1}) and GEDI trained without $\mathcal{L}_{NF}(\Theta)$ using two different encoders for computing the discriminative and the generative terms in our objective (called \textit{no NF, 2 enc.}). From the results in Figure~\ref{fig:ablation_toy}, we can make the following observations: (i) DAM plays a crucial role to recover the manifold structure, as the performance increase with the number of steps $T$ for \textit{GEDI}, \textit{no NF} and \textit{no NF, 2 enc.}. However, the strategy is not effective when we don't use the first training stage, as demonstrated by the results obtained by \textit{no NF, no train. 1} on circles. This confirms our original hypothesis that DAM requires a good generative model; (ii) When looking at the results obtained by \textit{no NF} and \textit{no NF, 2 enc.}, we observe that there is a clear advantage, especially on circles, by using a single encoder, highlighting the fact that the integration between generative and self-supervised models is effective not only at the objective level but also at the architectural one; (iii) From the comparison between \textit{GEDI} and \textit{no NF} on circles, we observe improved performance and reduced variance. This suggests that $\mathcal{L}_{NF}(\Theta)$ complements the other terms in our objective and therefore contributes to driving the learning towards desired solutions.

\begin{figure}
     \centering
     \begin{subfigure}[b]{0.45\linewidth}
         \centering
         \includegraphics[width=0.9\textwidth]{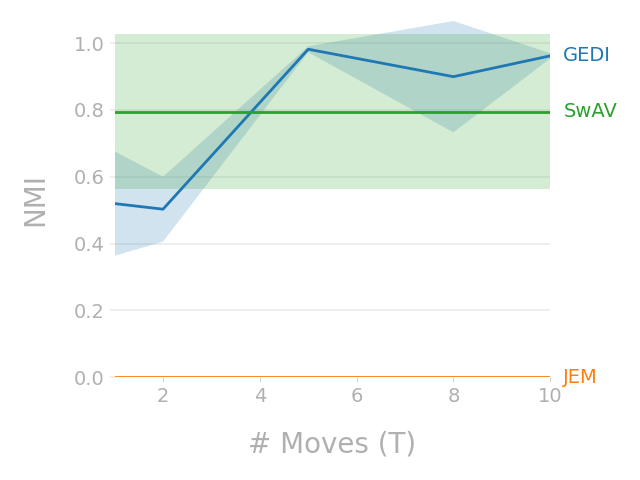}
         \caption{Moons}
     \end{subfigure}%
     \begin{subfigure}[b]{0.45\linewidth}
         \centering
         \includegraphics[width=0.9\textwidth]{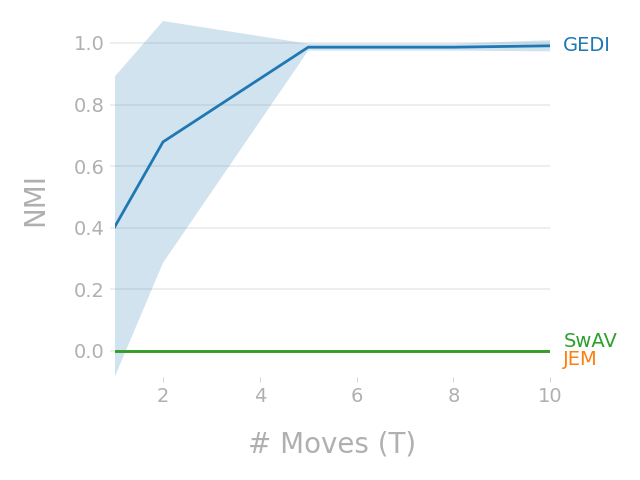}
         \caption{Circles}
     \end{subfigure}%     
     \caption{NMI achieved by JEM, SwAV and GEDI on moons and circles dataset. The curves are obtained by choosing different values of $T$ for DAM, namely $T\in\{1,2,5,8,10\}$.}
     \label{fig:nmi_toy}
\end{figure}

{\section{Hyperparameters for Synthetic Data}\label{sec:hyperparams_synth}}
For the backbone $enc$, we use a MLP with two hidden layers and 100 neurons per layer, an output layer with 2 neurons and LeakyReLU activation functions. For the projection head $proj$, we use a MLP with one hidden layer and 4 and an output layer with 2 neurons (batch normalization is used in all layers) and final $L_2$ normalization.
For JEM, we use an output layer with only one neuron. All methods use a batch size of 400.
Baseline JEM (following the original paper):
\begin{itemize}
    \item Number of iterations $100K$
    \item Learning rate $1e-3$
    \item Optimizer Adam $\beta_1=0$, $\beta_2=0.9$
    \item SGLD steps $1$
    \item Buffer size 10000
    \item Reinitialization frequency $0.05$
    \item SGLD step-size $1$
    \item SGLD noise $0.01$
\end{itemize}
And for self-supervised learning methods, please refer to Table~\ref{tab:hyperparams}.

{\section{Hyperparameters for SVHN, CIFAR-10, CIFAR-100}\label{sec:hyperparams_real}}
\begin{table}[ht]
  \caption{Resnet architecture. Conv2D(A,B,C) applies a 2d convolution to input with B channels and produces an output with C channels using stride (1, 1), padding (1, 1) and kernel size (A, A).}
  \label{tab:backbone}
  \centering
\begin{tabular}{@{}lrr@{}}
\toprule
\textbf{Name} & \textbf{Layer} & \textbf{Res. Layer} \\
\midrule
\multirow{6}{*}{Block 1} & Conv2D(3,3,F) & \multirow{2}{*}{AvgPool2D(2)} \\
& LeakyRELU(0.2) & \\
& Conv2D(3,F,F) & \multirow{2}{*}{Conv2D(1,3,F) no padding}\\
& AvgPool2D(2) & \\
& \cline{1-2}\\
& \multicolumn{2}{c}{Sum} \\
\midrule
\multirow{5}{*}{Block 2} & LeakyRELU(0.2) & \\
& Conv2D(3,F,F) & \\
& LeakyRELU(0.2) & \\
& Conv2D(3,F,F) & \\
& AvgPool2D(2) & \\
\midrule
\midrule
\multirow{4}{*}{Block 3} & LeakyRELU(0.2) & \\
& Conv2D(3,F,F) & \\
& LeakyRELU(0.2) & \\
& Conv2D(3,F,F) & \\
\midrule
\multirow{5}{*}{Block 4} & LeakyRELU(0.2) & \\
& Conv2D(3,F,F) & \\
& LeakyRELU(0.2) & \\
& Conv2D(3,F,F) & \\
& AvgPool2D(all) & \\
\bottomrule
\end{tabular}
\end{table}
For the backbone $enc$, we use a ResNet with 8 layers as in~\cite{duvenaud2021no}, where its architecture is shown in Table~\ref{tab:backbone}. For the projection head $proj$, we use a MLP with one hidden layer and $2*F$ neurons and an output layer with $F$ neurons (batch normalization is used in all layers) and final $L_2$ normalization. $F=128$ for SVHN, CIFAR-10 (1 million parameters) and $F=256$ for CIFAR-100 (4.1 million parameters).
 For JEM, we use the same settings of~\cite{duvenaud2021no}. All methods use a batch size of 64.
Baseline JEM (following the original paper):
\begin{itemize}
    \item Number of epochs $100$
    \item Learning rate $1e-4$
    \item Optimizer Adam
    \item SGLD steps $20$
    \item Buffer size 10000
    \item Reinitialization frequency $0.05$
    \item SGLD step-size $1$
    \item SGLD noise $0.01$
    \item Data augmentation (Gaussian noise) $0.03$
\end{itemize}
And for self-supervised learning methods, please refer to Table~\ref{tab:hyperparams}.
\begin{table*}[ht]

  \caption{Hyperparameters (in terms of sampling, optimizer, objective and data augmentation) used in all experiments.}
  \label{tab:hyperparams}
  \centering
\begin{tabular}{@{}llrrrr@{}}
\toprule
\textbf{Class} & \textbf{Name param.} & \textbf{SVHN} & \textbf{CIFAR-10} & \textbf{CIFAR-100} & \textbf{Addition} \\
\midrule
\multirow{5}{*}{Data augment.} & Color jitter prob. & 0.1 & 0.1 & 0.1 & 0.1 \\
& Gray scale prob. & 0.1 & 0.1 & 0.1 & 0.1 \\
& Random crop & Yes & Yes & Yes & Yes \\
& Additive Gauss. noise (std) & 0.03 & 0.03 & 0.03 & 0.3 \\
& Random horizontal flip & No & Yes & Yes & No \\
\midrule
            \multicolumn{6}{c}{Training 1} \\
\midrule
\multirow{5}{*}{SGLD} & SGLD iters & 20 & 20 & 20 & 10 \\
& Buffer size & 10k & 10k & 10k & 10k \\
& Reinit. frequency & 0.05 & 0.05 &0.05 & 0.05 \\
& SGLD step-size & 1 & 1 & 1 & 1 \\
& SGLD noise & 0.01 & 0.01 & 0.01 & 0.01 \\
\midrule
\multirow{6}{*}{Optimizer} & Batch size & 64 & 64 & 64 & 60 \\
& $\text{Iters}_1$ & 100k & 70k & 70k & Sec.~\ref{sec:hyperparams_nesy}\\
& Adam $\beta_1$ & 0.9 & 0.9 & 0.9 & 0.9 \\
& Adam $\beta_2$ & 0.999 & 0.999 & 0.999 & 0.999 \\
& Learning rate & $1e-4$ & $1e-4$ & $1e-4$ & $1e-4$ \\
\midrule
            \multicolumn{6}{c}{Training 2} \\
\midrule
\multirow{5}{*}{SGLD} & SGLD iters & idem & idem & idem & idem \\
& Buffer size & idem & idem & idem & idem \\
& Reinit. frequency & idem & idem & idem & idem \\
& SGLD step-size & idem & idem & idem & idem \\
& SGLD noise & idem & idem & idem & idem \\
\midrule
\multirow{6}{*}{Optimizer} & Batch size & idem & idem & idem & idem \\
& $\text{Iters}_2$ & idem & idem & idem & idem \\
& Adam $\beta_1$ & idem & idem & idem & idem \\
& Adam $\beta_2$ & idem & idem & idem & idem \\
& Learning rate & idem & idem & idem & idem \\
\midrule
\multirow{2}{*}{DAM} & $\epsilon$ & 0.03 & 0.03 & 0.03 & 0.03 \\
& $T$ & 10 & 10 & 10 & 10 \\
\midrule
\multirow{5}{*}{Weights for losses} & $-CE(p,p_\Psi)$ & 1 & 1 & 1 & 1 \\
&$\mathcal{L}_{NF}(\Theta)$ & 1/batch & 1/batch & 1/batch & 1/batch \\
& $\mathcal{L}_{DI}(\Theta,\mathcal{T})$ & 1000 & 1000 & 1000 & 1000 \\
& $\mathcal{L}_{DI}(\Theta,p_\Psi)$ & 500 & 500 & 500 & 500 \\
& $\mathcal{L}_{NeSY}(\Theta)$ & - & - & - & 0/3000 \\
\bottomrule
\end{tabular}
\end{table*}
\begin{table}[t]
  \caption{Ablation study for clustering performance in terms of normalized mutual information on test set (SVHN, CIFAR-10, CIFAR-100). Higher values indicate better clustering performance.}
  \label{tab:ablation_real}
  \centering
\begin{tabular}{@{}lrrrr@{}}
\toprule
\textbf{Dataset} & \textbf{no NF, 2 enc. (Ours)} & \textbf{no NF (Ours)} & \textbf{GEDI (Ourss)} & \textbf{Gain}  \\
\midrule
SVHN & 0.21 & 0.31 & \textbf{0.39} & \textbf{+0.08} \\
CIFAR-10 & 0.38 & \textbf{0.40} & \textbf{0.41} & +0.00\\
CIFAR-100 & \textbf{0.75} & \textbf{0.76} & 0.72s & -0.04 \\
\bottomrule
\end{tabular}
\end{table}
\begin{table*}
  \caption{Supervised linear evaluation in terms of accuracy on test set (SVHN, CIFAR-10, CIFAR-100). The linear classifier is trained for 100 epochs using SGD with momentum, learning rate $1e-3$ and batch size 100.}
  \label{tab:acc}
  \centering
  \begin{tabular}{@{}lllllll@{}}
    \toprule
    \textbf{Dataset} & \textbf{JEM} & \textbf{Barlow} & \textbf{SwAV} & \textbf{No NF, 2 enc.} & \textbf{No NF} & \textbf{GEDI} \\
    \midrule
    SVHN & 0.20 & 0.84 & 0.44 & 0.29 & 0.52 & 0.56 \\
    CIFAR-10 & 0.23 & 0.63 & 0.53 & 0.48 & 0.50 & 0.51 \\
    CIFAR-100 & 0.03 & 0.35 & 0.14 & 0.12 & 0.15 & 0.15 \\
    \bottomrule
  \end{tabular}
\end{table*}

\begin{table*}
  \caption{Generative performance in terms of Frechet Inception Distance (FID) (SVHN, CIFAR-10, CIFAR-100). The lower the values the better the performance are.}
  \label{tab:FID}
  \centering
  \begin{tabular}{@{}lllllll@{}}
    \toprule
    \textbf{Dataset} & \textbf{JEM} & \textbf{Barlow} & \textbf{SwAV} & \textbf{No NF, 2 enc.} & \textbf{No NF} & \textbf{GEDI} \\
    \midrule
    SVHN & 166 & 454 & 489 & \textbf{158} & 173 & 208 \\
    CIFAR-10 & 250 & 413 & 430 & \textbf{209} & 265 & 236 \\
    CIFAR-100 & 240 & 374 & 399 & \textbf{210} & 237 & 244 \\
    \bottomrule
  \end{tabular}
\end{table*}

\begin{table*}
  \caption{OOD detection in terms of AUROC on test set (CIFAR-10, CIFAR-100). Training is performed on SVHN.}
  \label{tab:oodsvhn}
  \centering
  \begin{tabular}{@{}lllllll@{}}
        \toprule
        \textbf{Dataset} & \textbf{JEM} & \textbf{Barlow} & \textbf{SwAV} & \textbf{No NF, 2 enc.} & \textbf{No NF} & \textbf{GEDI} \\
    \midrule
    CIFAR-10 & 0.75 & 0.43 & 0.21 & 0.76 & \textbf{0.97} & 0.94 \\
    CIFAR-100 & 0.75 & 0.5 & 0.28 & 0.75 & \textbf{0.94} & \textbf{0.93} \\
    \bottomrule
  \end{tabular}
\end{table*}

\begin{table*}
  \caption{OOD detection in terms of AUROC on test set (SVHN, CIFAR-100). Training is performed on CIFAR-10.}
  \label{tab:oodCIFAR-10}
  \centering
  \begin{tabular}{@{}lllllll@{}}
        \toprule
        \textbf{Dataset} & \textbf{JEM} & \textbf{Barlow} & \textbf{SwAV} & \textbf{No NF, 2 enc.} & \textbf{No NF} & \textbf{GEDI} \\
    \midrule
    SVHN & \textbf{0.43} & 0.31 & 0.24 & \textbf{0.43} & 0.31 & 0.31 \\
    CIFAR-100 & 0.54 & \textbf{0.56} & 0.51 & 0.53 & 0.53 & \textbf{0.55} \\
    \bottomrule
  \end{tabular}
\end{table*}

\begin{table*}
  \caption{OOD detection in terms of AUROC on test set (SVHN, CIFAR-10). Training is performed on CIFAR-100.}
  \label{tab:oodCIFAR-100}
  \centering
  \begin{tabular}{@{}lllllll@{}}
        \toprule
        \textbf{Dataset} & \textbf{JEM} & \textbf{Barlow} & \textbf{SwAV} & \textbf{No NF, 2 enc.} & \textbf{No NF} & \textbf{GEDI} \\
    \midrule
    SVHN & 0.48 & 0.43 & \textbf{0.50} & 0.47 & 0.32 & 0.38 \\
    CIFAR-10 & 0.48 & 0.42 & 0.46 & 0.47 & \textbf{0.49} & \textbf{0.50} \\
    \bottomrule
  \end{tabular}
\end{table*}
{\section{Ablation Study on SVHN, CIFAR-10, CIFAR-100}\label{sec:ablation}}
We conduct an ablation study to understand the impact of the different components of GEDI. We compare four different versions of GEDI, namely the full version (called simply \textit{GEDI}), GEDI trained without $\mathcal{L}_{NF}(\Theta)$ (called \textit{no NF}), GEDI trained without the first stage and also without $\mathcal{L}_{NF}(\Theta)$ (called \textit{no NF, no train. 1}) and GEDI trained without $\mathcal{L}_{NF}(\Theta)$ using two different encoders for computing the discriminative and the generative terms in our objective (called \textit{no NF, 2 enc.}). Results are shown in Table~\ref{tab:ablation_real}.

{\section{Additional Experiments on SVHN, CIFAR-10, CIFAR-100}\label{sec:additional}}
We conduct a linear probe evaluation of the representations learnt by the different models Table~\ref{tab:acc}. These experiments provide insights on the capabilities of learning representations producing linearly separable classes. From Table~\ref{tab:acc}, we observe a large difference in results between Barlow and SwAV. Our approach provides interpolating results between the two approaches.

We also evaluate the generative performance in terms of Frechet Inception Distance (FID). From Table~\ref{tab:FID}, we observe that GEDI outperforms all self-supervised baselines by a large margin, achieving comparable performance to JEM.

Additionally, we evaluate the performance in terms of OOD detection, by following the same methodology used in~\cite{grathwohl2020your}. We use the OOD score criterion proposed in~\cite{grathwohl2020your}, namely $s(x)=-\|\frac{\partial\log p_\Psi(x)}{\partial x}\|_2$. From Table~\ref{tab:oodsvhn}, we observe that GEDI achieves almost optimal performance. While these results are exciting, it is important to mention that they are not generally valid. Indeed, when training on CIFAR-10 and performing OOD evaluation on the other datasets, we observe that all approaches achieve similar performance both on CIFAR-100 and SVHN, suggesting that all datasets are considered in-distribution, see Table~\ref{tab:oodCIFAR-10}. A similar observation is obtained when training on CIFAR-100 and evaluating on CIFAR-10 and SVHN, see Table~\ref{tab:oodCIFAR-100}. Importantly, this is a phenomenon which has been only recently observed by the scientific community on generative models. Tackling this problem is currently out of the scope of this work. For further discussion about the issue, we point the reader to the works in~\cite{nalisnick2019deep}.

{\section{Details on the MNIST addition experiment.}\label{sec:hyperparams_nesy}}
We now discuss the details on the MNIST addition experiment.

\subsection{Hyperparameters}
For the backbone $enc$, we use a ResNet with 8 layers as in~\cite{duvenaud2021no}, where its architecture is shown in Table~\ref{tab:backbone}. For the projection head $proj$, we use a MLP with one hidden layer and 256 neurons and an output layer with 128 neurons (batch normalization is used in all layers) and final $L_2$ normalization. 
The number of epochs for both training phases for the three settings, i.e. $100$ examples, $1 000$ examples and $10 000$ examples are $100$, $30$ and $5$ epochs respectively. These were selected by the point at which the loss curves flatten out.

\subsection{Data generation}
The data was generated by uniformly sampling pairs $a,b$ such that $0 \le a \le 9$, $0 \le b \le 9$ and $0 \le a+b \le 9$. For each triplet $(a,b,c)$, we assigned to $a,b,c,$ random MNIST images with corresponding labels, without replacement.
\subsection{Calculating the constraint}
To calculate the constraint, we group the three images of each triplet consecutively in the batches, hence why the batch size is a multiple of 3. To calculate the probability of the constraint, we used an arithmetic circuit compiled from the DeepProbLog program that implements this constraint ~\cite{manhaeve2018deepproblog}.

%% file: src/contrastive.tex
Contrastive self-supervised learning can be interpreted in probabilistic terms using the graphical model in Fig.~\ref{fig:graph}(a). In particular, we can define the following conditional density:
\begin{equation}
p(\ell_i|x_i;\Theta)=\frac{e^{sim(g(x_{\ell_i}),g(x_i))/\tau}}{\sum_{j=1}^ne^{sim(g(x_j),g(x_i))/\tau}}
\label{eq:contrastive}
\end{equation}
where $sim$ is a similarity function, $\tau>0$ is temperature parameter used to calibrate the uncertainty for $p(\ell_i|x_i;\Theta)$ and $\Theta=\{\theta,\{x_i\}_i^n\}$ is the set of parameters, including the parameters of the embedding function and the observed data. The learning criterion can be obtained from the expected log-likelihood computed on the observed random quantities (and using the factorization provided by the graphical model), namely:
\begin{align}
    &\mathbb{E}_{p(x_{1:n},\ell_{1:n})}\big\{\log p(x_{1:n},\ell_{1:n};\Theta)\big\}\\
    &= \mathbb{E}_{\prod_{j=1}^np(x_j)\delta(\ell_j-j)}\big\{\log \prod_{i=1}^np(x_i)p(\ell_i|x_i;\Theta)\big\} \nonumber\\
    &=\sum_{i=1}^n \mathbb{E}_{p(x_i)}\big\{\log p(x_i)\big\}+\mathbb{E}_{\prod_{j=1}^np(x_j)\delta(\ell_j-j)}\bigg\{\sum_{i=1}^n\log p(\ell_i|x_i;\Theta)\bigg\} \nonumber\\
    &= \underbrace{\sum_{i=1}^n \mathbb{E}_{p(x_i)}\big\{\log p(x_i)\big\}}_\text{Negative entropy term, $-H_p(x_{1:n})$} + \underbrace{\mathbb{E}_{\prod_{j=1}^np(x_j)}\bigg\{\sum_{i=1}^n\log p(\ell_i=i|x_i;\Theta)\bigg\}}_\text{Conditional log-likelihood term $\mathcal{L}_{CT}(\Theta)$}
    \label{eq:contrastive_obj}
\end{align}
where $\delta$ in the second equality is a delta function and we use for instance $x_{1:n}$ as a compact way to express $x_1,\dots,x_n$. From Eq.~(\ref{eq:contrastive_obj}), we observe that the expected log-likelihood can be rewritten as the sum of two quantities, namely a negative entropy and a conditional log-likelihood terms. However, only the second addend in Eq.~(\ref{eq:contrastive_obj}) is relevant for maximization purposes over the parameters $\theta$. Importantly, we can now show that the conditional log-likelikood term in Eq.~(\ref{eq:contrastive_obj}), coupled with the definition provided in Eq.~(\ref{eq:contrastive}), corresponds to the notorious InfoNCE objective~\citep{oord2018representation}. To see this, let us recall InfoNCE:
\begin{align}
    \text{InfoNCE}\propto \mathbb{E}_{\prod_{j=1}^np(x_j,z_j)}\bigg\{\sum_{i=1}^n\log\frac{e^{f(x_i,x_i)}}{\sum_{k=1}^ne^{f(x_k,x_i)}}\bigg\} \nonumber
\end{align}
By choosing $p(x,z)=p(x)\delta(z-g(x))$ and $f(x,z)=sim(g(x),z)/\tau$ we recover the conditional log-likelihood term in Eq.~(\ref{eq:contrastive_obj}). Importantly, other contrastive objectives, such as CPC~\citep{henaff2020data}, SimCLR~\citep{chen2020simple}, ProtoCPC~\citep{lee2022prototypical}, KSCL~\citep{xu2022k} to name a few, can be obtained once we have a connection to InfoNCE.

%% file: src/discriminative.tex
Cluster-based SSL can be interpreted in probabilistic terms using the graphical model in Fig.~\ref{fig:graph}(b). In particular, we can define the following conditional density:
\begin{equation}
p(y_i|x_i;\Theta)=\frac{e^{U_{:y_i}^TG_{:i}/\tau}}{\sum_{y}e^{U_{:y}^TG_{:i}/\tau}}
\label{eq:discriminative}
\end{equation}
where $U\in\mathbb{R}^{h\times c}$ is a matrix\footnote{We use subscripts to select rows and columns. For instance, $U_{:y}$ identify $y-$th column of matrix $U$.} of $c$ cluster centers, $G=[g(x_1),\dots,g(x_n)]$ is a matrix of embeddings of size $h\times n$ and $\Theta=\{\theta,U\}$ is the set of parameters, including the ones for the embedding function and the cluster centers.  The learning criterion can be obtained in a similar way to what we have done previously for contrastive methods. In particular, we have that
\begin{align}
&\mathbb{E}_{p(x_{1:n})}\{\log p(x_{1:n};\Theta)\}\\
&=-H_p(x_{1:n})+ \mathbb{E}_{p(x_{1:n})\mathcal{T}(x_{1:n}'|x_{1:n})}\left\{\log\sum_{y_{1:n}}p(y_{1:n}|x_{1:n};\Theta)\right\}\nonumber\\
&=-H_p(x_{1:n})+ \mathbb{E}_{p(x_{1:n})\mathcal{T}(x_{1:n}'|x_{1:n})}\Big\{\log\sum_{y_{1:n}}\frac{q(y_{1:n}|x_{1:n}')}{q(y_{1:n}|x_{1:n}')} p(y_{1:n}|x_{1:n};\Theta)\} \nonumber\\
&\geq -H_p(x_{1:n})- \mathbb{E}_{p(x_{1:n})\mathcal{T}(x_{1:n}'|x_{1:n})}\{KL(q(y_{1:n}|x_{1:n}')\|p(y_{1:n}|x_{1:n}))\} \nonumber\\
&=\underbrace{\sum_{i=1}^n \mathbb{E}_{p(x_i)}\big\{\log p(x_i)\big\}}_\text{Negative entropy term, $-H_p(x_{1:n})$}\nonumber\\
&\quad+\underbrace{\sum_{i=1}^n\mathbb{E}_{p(x_i)\mathcal{T}(x_i'|x_i)}\{\mathbb{E}_{q(y_i|x_i')}\log p(y_i|x_i;\Theta) + H_q(y_i|x_i')\}}_\text{Discriminative term $\mathcal{L}_{DI}(\Theta)$}
\label{eq:discriminative_obj}
\end{align}
where $q(y_i|x_i')$ is an auxiliary distribution. Notably, maximizing the discriminative term is equivalent to minimize the $KL$ divergence between the two predictive distributions $p(y_i|x_i)$ and $q(y_i|x_i')$, thus learning to predict similar category for both sample $x_i$ and its augmented version $x_i'$, obtained through $\mathcal{T}$. Importantly, we can relate the criterion in Eq.~(\ref{eq:discriminative_obj}) to the objective 
used in optimal transport~\cite{cuturi2013sinkhorn}, by substituting Eq.~(\ref{eq:discriminative}) into Eq.~(\ref{eq:discriminative_obj}) and adopting a matrix format, namely:
\begin{align}
    \mathcal{L}_{DI}(\Theta;\mathcal{T}) &=\frac{1}{\tau}\bigg\{ \mathbb{E}_{p(x_{1:n})\mathcal{T}(x_i'|x_i)}\{Tr(QU^TG)\}+\tau\mathbb{E}_{p(x_{1:n})\mathcal{T}(x_i'|x_i)}\{H_Q(y_{1:n}|x_{1:n}')\}\bigg\} 
    \label{eq:discriminative_obj2}
\end{align}
where $Q=[q(y_1|x_1'),\dots,q(y_n|x_n')]^T$ is a prediction matrix of size $n\times c$ and $Tr(A)$ is the trace of an arbitrary matrix $A$. Note that a naive maximization of $\mathcal{L}_{DI}(\Theta)$ can lead to obtain trivial solutions like 
the one corresponding to uniformative predictions, namely $q(y_i|x_i')=p_\gamma(y_i|x_i)=\text{Uniform}(\{1,\dots,c\})$ for all $i=1,\dots,n$. Fortunately, the problem can be avoided and solved exactly using the Sinkhorn-Knopp algorithm, which alternates between maximizing $\mathcal{L}_{DI}(\Theta)$ in Eq.~(\ref{eq:discriminative_obj2}) with respect to $Q$ and with respect to $\Theta$, respectively. This is indeed the procedure used in several cluster-based SSL approaches, like DeepCluster~\cite{caron2018deep} and SwAV~\cite{caron2020unsupervised}, to name a few.

%% file: src/negativefree.tex
We can provide a probabilistic interpretation also for negative-free SSL using the graphical model shown in Fig.~\ref{fig:graph}(c). Note that, for the sake of simplicity in the graph and in the following derivation, the latent variables ($w$ and all $\xi_i$) are considered independent of each other only for the generation process. In reality, one should consider an alternative but equivalent model, using a generating process including also the edges $x_i\rightarrow w$, $\xi_i\rightarrow x_i'$ and defining $p(w|x_i)=p(w)$ and $p(x_i'|x_i,\xi_i)=\mathcal{T}(x_i'|x_i)$ for all $i=1\dots,n$. Based on these considerations, we can define the prior and our auxiliary and inference densities for the model in the following way:
\begin{align}
    p(w) &= \mathcal{N}(w|0,I) \nonumber\\
    p(\xi_i) &= \mathcal{N}(\xi_i|0,I) \nonumber\\
    q(w|x_{1:n};\Theta) &=\mathcal{N}(w|0,\Sigma) \nonumber\\
    q(\xi_i|x_i,x_i';\Theta) &= \mathcal{N}(\xi_i|enc(x_i)-enc(x_i'),I)
    \label{eq:negativefree}
\end{align}
where $\mathcal{N}(\cdot|\mu,\Sigma)$ refers to a multivariate Gaussian density with mean $\mu$ and covariance $\Sigma$, $I$ is an identity matrix, $\Sigma=\sum_{i=1}^n(g(x_i)-\bar{g})(g(x_i)-\bar{g})^T+\beta I$, $\beta$ is positive scalar used to ensure the positive-definiteness of $\Sigma$, $\bar{g}=1/n\sum_{i=1}^ng(x_i)$ and $\Theta=\{\theta\}$.
Importantly, while $q(w|x_{1:n})$ in Eq.~(\ref{eq:negativefree}) is used to store the global statistical information of the data in the form of an unnormalized sample covariance, $q(\xi_i|x_i,x_i')$ is used to quantify the difference between a sample and its augmented version in terms of their latent representation.
Similarly to previous SSL classes and by reusing definitions in Eq.~(\ref{eq:negativefree}), we can devise the learning criterion in the following way:
\begin{align}
\mathbb{E}_{p(x_{1:n})}\{\log p(x_{1:n};\Theta)\}&\geq -H_p(x_{1:n}) - \mathbb{E}_{p(x_{1:n})}\{KL(q(w|x_{1:n};\Theta)\|p(w))\} \nonumber\\
& \quad -\mathbb{E}_{p(x_{1:n})\mathcal{T}(x_{1:n}'|x_{1:n})}\{KL(q(\xi_{1:n}|x_{1:n},x_{1:n}';\Theta)\|p(\xi_{1:n}))\}\nonumber\\
&= -H_p(x_{1:n}) - \mathbb{E}_{p(x_{1:n})}\{KL(q(w|x_{1:n};\Theta)\|p(w))\} \nonumber\\
& \quad -\sum_{i=1}^n\mathbb{E}_{p(x_i)\mathcal{T}(x_i'|x_i)}\{KL(q(\xi_i|x_i,x_i';\Theta)\|p(\xi_i))\}\nonumber\\
&\propto\underbrace{\sum_{i=1}^n \mathbb{E}_{p(x_i)}\big\{\log p(x_i)\big\}}_\text{Negative entropy term, $-H_p(x_{1:n})$}\underbrace{-\mathbb{E}_{p(x_{1:n})}\left\{\frac{Tr(\Sigma)}{2}{-} \frac{\log|\Sigma|}{2}\right\}}_\text{Negative-free term, $\mathcal{L}_{NF}(\Theta)$ (first part)}\nonumber\\
& \qquad\underbrace{-\sum_{i=1}^n\mathbb{E}_{p(x_i)\mathcal{T}(x_i'|x_i)}\left\{\frac{dist(x_i,x_i')}{2}\right\}}_\text{Negative-free term, $\mathcal{L}_{NF}(\Theta)$ (second part)}
\label{eq:negativefree_obj}
\end{align}
where $dist(x,x')=\|enc(x)-enc(x')\|^2$ and $|A|$ computes the determinant of an arbitrary matrix $A$. Notably, the maximization of $\mathcal{L}_{NF}(\Theta)$ promotes both decorrelated embedding features, as the first two addends in $\mathcal{L}_{NF}(\Theta)$ (obtained from the first KL term in Eq.~(\ref{eq:negativefree_obj})) force $\Sigma$ to become an identity matrix, as well as representations that are invariant to data augmentations, thanks to the third addend in $\mathcal{L}_{NF}(\Theta)$. It is important to mention that $\mathcal{L}_{NF}(\Theta)$ in Eq.~(\ref{eq:negativefree_obj}) recovers two recent negative-free criteria, namely CorInfoMax~\citep{ozsoy2022self} and MEC~\citep{liu2022self}. We can also relate $\mathcal{L}_{NF}(\Theta)$ to other existing negative-free approaches, including Barlow Twins~\citep{zbontar2021barlow}, VicReg~\citep{bardes2022vicreg,bardes2022vicregl} and W-MSE~\citep{ermolov2021whitening}.

%ELBOgen
%EBM and negative-free methods. Recently, there
%has been a surge of interest for approaches that do not rely
%on negative samples [2], [3], [4], [5], [6]. This allows to re-
%duce the computational burden of contrastive-like methods,
%which usually operate with large batch sizes. We can also
%view negative-free objectives from a likelihood perspective
%and highlight the main underlying assumptions. Indeed, for
%a uniform prior p(z), we have that Eq. (1) reduces to the
%following objective:
%ELBOgen
%EBM = Ep(x)q(z|x){fθ (x, z)} − Ep(x)q(z|x){log Γ(z; θ)} + Ep(x)q(z
%(4)
%where η is a constant equal to log p(z). Importantly, recent
%negative-free methods, like Barlow Twins [5], W-MSE [12]
%and VICReg [6], can be reduced to the optimization of the
%objective function in Eq. (4) without the second addend
%(please refer to Appendix E for further technical details).
%Consequently, these approaches learn an unnormalized en-
%ergy model and, as we will see in the experiments, this
%observation has important consequences on their perfor-
%mance.

%% file: src/unified.tex
In all three classes of SSL approaches (see Eqs.~(\ref{eq:contrastive_obj}),(\ref{eq:discriminative_obj}) and (\ref{eq:negativefree_obj})), the expected data log-likelihood can be lower bounded by the sum of two contribution terms, namely a negative entropy $-H_p(x_{1:n})$ and a conditional log-likelihood term, chosen from $\mathcal{L}_{CT}(\Theta),\mathcal{L}_{DI}(\Theta)$ and $\mathcal{L}_{NF}(\Theta)$. A connection to generative models emerges by additionally lower bounding the negative entropy term, namely:
\begin{align}
    -H_p(x_{1:n})&=\mathbb{E}_{p(x_{1:n})}\{\log p(x_{1:n})\} \nonumber\\
    &=\sum_{i=1}^n\mathbb{E}_{p(x_i)}\{\log p(x_i)\} \nonumber\\
    &=\sum_{i=1}^n\left[\mathbb{E}_{p(x_i)}\{\log p_\Psi(x_i)\}+KL(p(x_i)\|p_\Psi(x_i))\right] \nonumber\\
    &\geq \underbrace{\sum_{i=1}^n\mathbb{E}_{p(x_i)}\{\log p_\Psi(x_i)\}}_\text{$-CE(p,p_\Psi)$}
    \label{eq:generative_obj}
\end{align}
where $p_\Psi(x)$ is a generative model parameterized by $\Psi$. Notably, the relation in~(\ref{eq:generative_obj}) can be substituted in any of the objectives previously derived for the different SSL classes, thus allowing to jointly learn both generative and SSL models. This leads to a new GEnerative and DIscriminative family of models, which we call GEDI. Importantly, any kind of likelihood-based generative model (for instance variational autoencoders, normalizing flows, autoregressive or energy-based models) can be considered in GEDI. In this work, we argue that much can be gained by leveraging the GEDI integration. Notably, there has been a recent work EBCLR~\citep{kim2022energy} integrating energy-based  models with contrastive SSL approaches. Here, we show that EBCLR represents one possible instantiation of GEDI. For instance, let us consider contrastive SSL and observe that the conditional density in Eq.~(\ref{eq:contrastive}) can be decomposed into a joint and a marginal densities (similarly to what is done in~\citep{grathwohl2020your}):
\begin{align}
    p(\ell,x;\Theta) &= \frac{e^{sim(g(x_{\ell}),g(x))/\tau}}{\Gamma(\Theta)} \nonumber\\
    p(x;\Theta) &= \frac{\sum_{j=1}^ne^{sim(g(x_{\ell_j}),g(x))/\tau}}{\Gamma(\Theta)} \nonumber\\
    &= \frac{e^{-\underbrace{\left(-\log\sum_{j=1}^ne^{sim(g(x_{\ell_j}),g(x))/\tau}\right)}_\text{$E(x;\Theta)$}}}{\Gamma(\Theta)}
    \label{eq:generative_ebm}
\end{align}
where $E(x,\Theta)$ defines the energy score of the marginal density. Now, by choosing $p_{\Psi}(x)=p(x;\Theta)$ and $sim(z,z')=-\|z-z'\|^2$ in Eq.~(\ref{eq:generative_ebm}), one recovers the exact formulation of EBCLR~\citep{kim2022energy}.

%% file: gedi.bbl
\begin{thebibliography}{27}
\providecommand{\natexlab}[1]{#1}
\providecommand{\url}[1]{\texttt{#1}}
\expandafter\ifx\csname urlstyle\endcsname\relax
  \providecommand{\doi}[1]{doi: #1}\else
  \providecommand{\doi}{doi: \begingroup \urlstyle{rm}\Url}\fi

\bibitem[Bardes et~al.(2022{\natexlab{a}})Bardes, Ponce, and
  Lecun]{bardes2022vicreg}
A.~Bardes, J.~Ponce, and Y.~Lecun.
\newblock {VICReg: Variance-Invariance-Covariance Regularization for
  Self-Supervised Learning}.
\newblock In \emph{ICLR}, 2022{\natexlab{a}}.

\bibitem[Bardes et~al.(2022{\natexlab{b}})Bardes, Ponce, and
  LeCun]{bardes2022vicregl}
A.~Bardes, J.~Ponce, and Y.~LeCun.
\newblock {VICREGL: Self-Supervised Learning of Local Visual Features}.
\newblock In \emph{NeurIPS}, 2022{\natexlab{b}}.

\bibitem[Caron et~al.(2018)Caron, Bojanowski, Joulin, and Douze]{caron2018deep}
M.~Caron, P.~Bojanowski, A.~Joulin, and M.~Douze.
\newblock {Deep Clustering for Unsupervised Learning of Visual Features}.
\newblock In \emph{ECCV}, 2018.

\bibitem[Caron et~al.(2020)Caron, Misra, Mairal, Goyal, Bojanowski, and
  Joulin]{caron2020unsupervised}
M.~Caron, I.~Misra, J.~Mairal, P.~Goyal, P.~Bojanowski, and A.~Joulin.
\newblock {Unsupervised Learning of Visual Features by Contrasting Cluster
  Assignments}.
\newblock In \emph{NeurIPS}, 2020.

\bibitem[Chen et~al.(2020)Chen, Kornblith, Norouzi, and Hinton]{chen2020simple}
T.~Chen, S.~Kornblith, M.~Norouzi, and G.~Hinton.
\newblock {A Simple Framework for Contrastive Learning of Visual
  Representations}.
\newblock In \emph{ICML}, 2020.

\bibitem[Cuturi(2013)]{cuturi2013sinkhorn}
M.~Cuturi.
\newblock {Sinkhorn Distances: Lightspeed Computation of Optimal Transport}.
\newblock In \emph{NeurIPS}, 2013.

\bibitem[da~Costa et~al.(2022)da~Costa, Fini, Nabi, Sebe, and
  Ricci]{costa2022solo}
V.~G.~T. da~Costa, E.~Fini, M.~Nabi, N~Sebe, and E.~Ricci.
\newblock {Solo-learn: A Library of Self-supervised Methods for Visual
  Representation Learning}.
\newblock \emph{JMLR}, 23\penalty0 (56):\penalty0 1--6, 2022.

\bibitem[den Oord et~al.(2018)den Oord, Li, and
  Vinyals]{oord2018representation}
A.~Van den Oord, Y.~Li, and O.~Vinyals.
\newblock {Representation Learning with Contrastive Predictive Coding}.
\newblock In \emph{arXiv}, 2018.

\bibitem[Du \& Mordatch(2019)Du and Mordatch]{du2019implicit}
Y.~Du and I.~Mordatch.
\newblock {Implicit Generation and Generalization in Energy-Based Models}.
\newblock \emph{arXiv}, 2019.

\bibitem[Duvenaud et~al.(2021)Duvenaud, Kelly, Swersky, Hashemi, Norouzi, and
  Grathwohl]{duvenaud2021no}
D.~Duvenaud, J.~Kelly, K.~Swersky, M.~Hashemi, M.~Norouzi, and W.~Grathwohl.
\newblock {No MCMC for Me: Amortized Samplers for Fast and Stable Training of
  Energy-Based Models}.
\newblock In \emph{ICLR}, 2021.

\bibitem[Ermolov et~al.(2021)Ermolov, Siarohin, Sangineto, and
  Sebe]{ermolov2021whitening}
A.~Ermolov, A.~Siarohin, E.~Sangineto, and N.~Sebe.
\newblock {Whitening for Self-Supervised Representation Learning}.
\newblock In \emph{ICML}, 2021.

\bibitem[Grathwohl et~al.(2020)Grathwohl, Wang, Jacobsen, Duvenaud, Norouzi,
  and Swersky]{grathwohl2020your}
W.~Grathwohl, K.-C. Wang, J.-H. Jacobsen, D.~Duvenaud, M.~Norouzi, and
  K.~Swersky.
\newblock {Your Classifier is Secretly an Energy Based Model and You Should
  Treat It Like One}.
\newblock In \emph{ICLR}, 2020.

\bibitem[Kim \& Ye(2022)Kim and Ye]{kim2022energy}
B.~Kim and J.~C. Ye.
\newblock {Energy-Based Contrastive Learning of Visual Representations}.
\newblock In \emph{NeurIPS}, 2022.

\bibitem[Lee(2022)]{lee2022prototypical}
K.~Lee.
\newblock {Prototypical Contrastive Predictive Coding}.
\newblock In \emph{ICLR}, 2022.

\bibitem[Liu et~al.(2022)Liu, Wang, Li, and Wang]{liu2022self}
X.~Liu, Z.~Wang, Y.~Li, and S.~Wang.
\newblock {Self-Supervised Learning via Maximum Entropy Coding}.
\newblock In \emph{NeurIPS}, 2022.

\bibitem[Manhaeve et~al.(2018)Manhaeve, Dumancic, Kimmig, Demeester, and
  Raedt]{manhaeve2018deepproblog}
R.~Manhaeve, S.~Dumancic, A.~Kimmig, T.~Demeester, and L.~De Raedt.
\newblock {DeepProbLog: Neural Probabilistic Logic Programming}.
\newblock In \emph{NeurIPS}, 2018.

\bibitem[Nalisnick et~al.(2019)Nalisnick, Matsukawa, Teh, Gorur, and
  Lakshminarayanan]{nalisnick2019deep}
E.~Nalisnick, A.~Matsukawa, Y.~W. Teh, D.~Gorur, and B.~Lakshminarayanan.
\newblock Do deep generative models know what they don't know?
\newblock In \emph{ICLR}, 2019.

\bibitem[Nijkamp et~al.(2019)Nijkamp, Hill, Zhu, and Wu]{nijkamp2019learning}
E.~Nijkamp, M.~Hill, S.~C. Zhu, and Y.~N. Wu.
\newblock Learning non-convergent non-persistent short-run mcmc toward
  energy-based model.
\newblock In \emph{NeurIPS}, 2019.

\bibitem[Nijkamp et~al.(2020)Nijkamp, Hill, Han, Zhu, and
  Wu]{nijkamp2020anatomy}
E.~Nijkamp, M.~Hill, T.~Han, S.-C. Zhu, and Y.~N. Wu.
\newblock {On the Anatomy of MCMC-Based Maximum Likelihood Learning of
  Energy-Based Models}.
\newblock In \emph{AAAI}, 2020.

\bibitem[O.~Henaff(2020)]{henaff2020data}
Olivier O.~Henaff.
\newblock {Data-Efficient Image Recognition with Contrastive Predictive
  Coding}.
\newblock In \emph{ICML}, 2020.

\bibitem[O.Chapelle et~al.(2010)O.Chapelle, Sch{\"o}lkopf, and
  Zien]{chapelle2010semi}
O.Chapelle, B.~Sch{\"o}lkopf, and A.~Zien.
\newblock {Semi-Supervised Learning. Adaptive Computation and Machine
  Learning}.
\newblock \emph{Methods}, 2010.

\bibitem[Ozsoy et~al.(2022)Ozsoy, Hamdan, Arik, Yuret, and
  Erdogan]{ozsoy2022self}
S.~Ozsoy, S.~Hamdan, S.~Arik, D.~Yuret, and A.~T. Erdogan.
\newblock {Self-Supervised Learning with an Information Maximization
  Criterion}.
\newblock In \emph{NeurIPS}, 2022.

\bibitem[Sansone \& Manhaeve(2022)Sansone and Manhaeve]{sansone2022gedi}
E.~Sansone and R.~Manhaeve.
\newblock {GEDI: GEnerative and DIscriminative Training for Self-Supervised
  Learning}.
\newblock In \emph{arXiv}, 2022.

\bibitem[Xie et~al.(2016)Xie, Lu, Zhu, and Wu]{xie2016theory}
J.~Xie, Y.~Lu, S.~C. Zhu, and Y.~N. Wu.
\newblock {A Theory of Generative Convnet}.
\newblock In \emph{ICML}, 2016.

\bibitem[Xu et~al.(2022)Xu, Xiong, and Qi]{xu2022k}
H.~Xu, H.~Xiong, and G.~J. Qi.
\newblock {K-Shot Contrastive Learning of Visual Features with Multiple
  Instance Augmentations}.
\newblock \emph{IEEE Transactions on Pattern Analysis and Machine
  Intelligence}, 2022.

\bibitem[Xu et~al.(2018)Xu, Zhang, Friedman, Liang, and Broeck]{xu2018semantic}
J.~Xu, Z.~Zhang, T.~Friedman, Y.~Liang, and G.~Broeck.
\newblock {A Semantic Loss Function for Deep Learning with Symbolic Knowledge}.
\newblock In \emph{ICML}, 2018.

\bibitem[Zbontar et~al.(2021)Zbontar, Jing, Misra, LeCun, and
  Deny]{zbontar2021barlow}
J.~Zbontar, L.~Jing, I.~Misra, Y.~LeCun, and S.~Deny.
\newblock {Barlow Twins: Self-Supervised Learning via Redundancy Reduction}.
\newblock In \emph{ICML}, 2021.

\end{thebibliography}
